\documentclass{article}

\usepackage[a4paper, margin=2.5cm]{geometry}

\usepackage{amsmath,amsfonts}
\usepackage{algorithmic}
\usepackage{algorithm}
\usepackage{array}
\usepackage{textcomp}
\usepackage{stfloats}
\usepackage{xurl}
\usepackage{verbatim}
\usepackage{cite}
\usepackage{siunitx}
\usepackage{multirow}
\usepackage{enumitem}
\usepackage{graphicx}
\usepackage{amssymb}  
\usepackage{caption} 
\usepackage{epstopdf}  

\usepackage{balance}
\usepackage{soul}
\newcommand{\ch}[1]{\textcolor{black}{#1}}
\usepackage{fancyhdr}  
\title{\LARGE \bf Propeller damage detection, classification and estimation in multirotor vehicles}

\author{Claudio Pose*, Juan Giribet, and Gabriel Torre
    \thanks{Claudio Pose is with Laboratorio de Automática y Robótica, Facultad de Ingeniería, Universidad de Buenos Aires and CONICET - Universidad de San Andrés {\tt\small cldpose@fi.uba.ar}}
    \thanks{Juan Giribet is with Laboratorio de Inteligencia Artificial y Robótica, Universidad de San Andrés and CONICET {\tt\small jgiribet@conicet.gov.ar}}
    \thanks{Gabriel Torre is with Laboratorio de Inteligencia Artificial y Robótica, Universidad de San Andrés and Instituto de Ingeniería Biomédica, Facultad de Ingeniería, Universidad de Buenos Aires {\tt\small torreg@udesa.edu.ar}}
}

\date{}

\begin{document}

\maketitle

\begin{abstract}
This manuscript details an architecture and training methodology for a data-driven framework aimed at detecting, identifying, and quantifying damage in the propeller blades of multirotor Unmanned Aerial Vehicles. By substituting one propeller with a damaged counterpart—encompassing three distinct damage types of varying severity—real flight data was collected. This data was then used to train a composite model, comprising both classifiers and neural networks, capable of accurately identifying the type of failure, estimating damage severity, and pinpointing the affected rotor. The data employed for this analysis was exclusively sourced from inertial measurements and control command inputs, ensuring adaptability across diverse multirotor vehicle platforms.
\end{abstract}

\section{Introduction}
\subsection{Emergence and Importance of Multirotor UAVs}

Over the last decade, multirotor unmanned aerial vehicles (UAVs) surged in popularity due to their simplicity, affordable cost, and ease of maintenance. This led to an increase in UAV-dependent professional in diverse fields such as agriculture, package delivery, and  aerial photography.

As UAV usage grows, particularly in close proximity to humans, the safety of these vehicles has become paramount, requiring a high level of reliability from the aircraft. The operator (or team of operators) must always ensure good maintenance of the different parts of the vehicle to operate at the highest possible safety levels. However, UAVs remain susceptible to environmental damage during operation.

\subsection{Challenges in UAV Maintenance and Damage Detection}

\ch{
In this aspect, one of the parts of the vehicle that generally demands more maintenance and whose failure is critical to ensure the continuity of a flight is the actuator subsystem, composed of a given number of rotor-propeller pairs. Being the primary moving part in most multirotors, barring certain exceptional designs, this subsystem is highly susceptible to wear and tear. Factors such as harsh environmental conditions (e.g., operations in dusty or sandy areas) and physical impacts can significantly degrade its performance. This degradation reduces efficiency and increases unwanted vibrations, potentially leading to catastrophic failures and crashes. 
\\
While some types of damage can be precisely modeled through physical equations, for other types, it may be impossible to do so. For example, dents, broken tips, or structural damage in propellers may behave in different ways according to their rotational speed and acceleration, and to the state of the environment, considering their interaction with the wind and other factors. These complexities make it difficult, if not impossible, to precisely model their impact using traditional physical equations. This may severely restrict the use of model-based control approaches, which are commonly used for maintaining UAV performance in varying conditions. These methods rely on accurate system dynamics models, which must be continuously refined to adapt to changes in the operating environment. Still, self-supervised learning techniques have been developed to enhance this adaptive process, enabling real-time inference and control even in operating regimes that differ significantly from the training data \cite{Saviolo2024}. 
\\
However, recent advances in learning-based methods, particularly Reinforcement Learning (RL), have shown promise in addressing such complexities by improving the control and adaptation of UAVs under uncertain conditions \cite{Eschmann2023}. RL-based approaches, while powerful, may require substantial computational resources and may face challenges in bridging the simulation-to-reality gap, which is critical in real-time applications.
\\
While non-critical damage may be repaired post-flight during maintenance routines, permanently remote-operated systems (which cannot be accessed or repaired but are still able to fly) benefit from precise knowledge of the nature and degree of failure for mission planning. 
By integrating these advanced control strategies, trajectory planning can be optimized to reduce stress on failing components, resulting in safer and more energy-efficient flights. Moreover, these advanced control strategies can contribute to preventive maintenance by identifying potential failures before they occur, thus enabling higher-level decision-making processes. By anticipating issues, operators can plan maintenance interventions more effectively and adjust mission parameters to mitigate the impact of potential failures. This proactive approach not only enhances the safety and reliability of UAV operations but also optimizes resource usage and mission outcomes.}

\subsection{Recent Developments}

In recent years, data-based solutions, particularly deep neural networks, have made significant progress in efficiently solving various tasks, often simplifying the algorithm discovery process. While not a one-tool-for-all-problems solution, these approaches can offer valuable insights, especially in cases like the one presented here, where the observed measurement does not exhibit a clear relation with the issue that needs to be identified or estimated. However, it is crucial to have a substantial amount of data to perform the training task effectively. Such solutions have seen increasing use in addressing the challenges of fault detection, identification and isolation in unmanned aerial vehicles, and have even been employed in proposing fault recovery algorithms.

The thrust and torque generated by the actuators (comprising rotor, propeller, and speed controller) in a multirotor UAV work together to lift the vehicle, provide stability, and enable controlled movements in different axes for navigation. For this reason, total failures in a UAV actuator (motor speed controller, rotor, and/or propeller) are easier to detect and isolate, as the lack of any actuator results in a predictable dynamic over a finite set. This is usually solved by means of an observer-based solution, as presented in \cite{Alwi2013,Alwi2014,Vey2016b,Saied2015a}, where different techniques based in residues, sliding mode, and nonlinear models are presented, and for which a rather precise model of the vehicle is needed. A comprehensive survey on the topic can be found in \cite{Fourlas2021}, along with a number of fault detection techniques for this kind of vehicles. Data-based solutions for the same problem have also been proposed, as in \cite{Dutta2019b,Dutta2021,Etchemaite2023}, where accurate and fast detections may be reached without the need of a physical model.

However, when the type of failures becomes more difficult to discern by means of a predictable behavior of a model, and even more when the relationship between the nature of the failure and the effects on the vehicle are difficult to model or even know if there is a relationship, data-based approaches become a useful tool. In the case of \cite{Baskaya2017}, a Support Vector Machine (SVM) classifier is proposed to deal with loss of efficiency in a propeller, which is a fairly common problem in multirotor UAVs due to wear of the actuators and harsh environmental conditions. Another SVM classifier is proposed in \cite{Saied2017}, in order to detect failures in a coaxial octorotors using current and speed direct measurements, where the presence of two contiguous rotors makes difficult to know which one presents the failure. While the actuator set as a whole has been focused by several researchers, the most common failures addressed have been generally related to loss of efficiency, total failure, or blockage.

Lately, as new techniques develop, interest has risen into a specific category of failures, namely those that relate to the interaction between rotor and propeller, and particularly structural damage in the latter. For example, \cite{Kantue2020} proposes a neural network approach to detect a slip between the rotor and propeller, for which it is necessary to measure the speed of the propellers. A similar problem is discussed in \cite{Sadhu2023}, where the solution is based on Deep Convolutional and Long Short Term Memory Neural Networks (Deep CNN and LSTM) to detect propeller breakage. 

Structural damage in propellers are of particular interest, as they need to be in perfect conditions, or its high-speed rotating unbalanced mass can inject perturbations into the structure and affect several sensors, or even deteriorate the rest of the structure. Inertial sensors, which are the most basic sensors a UAV needs to fly, are severely affected by vibrations in the vehicle's structure, to the point that the information they provide may be rendered unusable. However, these sensors are also useful to detect the occurrence of such damage, and also to establish some relationship between the nature of the damage and the perturbation it produces. The works in \cite{Ghalamchi2018,Ghalamchi2020} show that, if working with the frequency domain of the inertial data, it is possible to detect the frequency of rotation of the motors, as they produce small but noticeable vibrations around that frequency. Then, as damage that affects a propeller appears as a noticeable vibration in the frequency domain, the displacement in frequency or a change in the power density can be used to detect the nature and characteristics of such damage. Several works have exploited the frequency spectrum properties, such as \cite{Benini2019}, where this information is used to detect the presence of a damaged propeller, limited to the take-off and landing of the vehicle. This solution is also implemented in \cite{Bondyra2018}, where inertial sensors are added next to each rotor, in order to capture the perturbation injected by the damaged propellers, and using the energy in several frequency bands to detect in which motor the damage was present, including simultaneous damage in two propellers. The method is extensible to other sensors that provide similar information, such as acoustic sensors, 
which can also provide data in a larger spectrum.
Solutions using microphones arrays have been proposed in \cite{Iannace2019,Bondyra2022}. An interesting approach that avoids using the Fast Fourier Transform to obtain information about the frequency spectrum is proposed in \cite{Zhang2022}, with an application to detection of partial damage in a propeller.

Some researchers have provided several datasets for the community, with diverse data logged for some particular failures, as the nature of the possible damage types and degree of damage is infinite. The works in \cite{Gururajan2019,Baldini2023}, provide recorded flight data of several inertial sensors in a hexarotor with a damaged carbon fiber propeller, and of a commercial flight controller with additional speed sensors for each rotor in a hexarotor with a chipped plastic propeller, respectively. As the process of gathering such data not only requires countless hours of flying, but also the destruction of several propellers (or other components) and the possibility of several crashes, this information is very valuable, as even if the type of failure in different datasets is the same, it provides the opportunity to evaluate its effects on different vehicles.

Existing research on propeller blade damage often encounters one or more of these challenges:
\begin{enumerate}[label=\textbf{\alph*)}]
    \item Additional sensors other than those strictly necessary to fly a multirotor UAV are needed for the proposed solution (speed, current, acoustic, additional inertial sensors), as seen in \cite{Saied2017, Iannace2019, Kantue2020, Bondyra2018, Bondyra2022}.
    \item There is a need for sensors that may be common for UAV, but they don't work in every possible environment (GPS or other positioning systems, optical flow).
    \item Only a particular type of damage is taken into account (total failure, loss of efficiency, chipped blade, or imbalance), as considered in \cite{Baskaya2017, Iannace2019, Dutta2021, Etchemaite2023}.
    \item Only the existence of the damage is detected, but not its nature or location, or is limited to a small number of possible failures, as in \cite{Iannace2019, Sadhu2023}.    
\end{enumerate}

\subsection{Study Objectives}

In this work, the objective is to provide a data-based architecture which is able to detect the existence or occurrence of damage in a propeller, detect the location of the damaged propeller, and estimate the type and degree of damage it presents. We consider several types of damage, each potentially causing one or more effects noted in prior research, including loss of efficiency, mass imbalance, and changes in effective area. Such an architecture will provide a solution that allows for precise and proactive maintenance, and also allow for in-flight fault recovery. A custom dataset will be acquired, and then the proposed solution will be evaluated, using a commercially available quadrotor platform.

This work will use as input data only that which is available in any kind of multirotor aerial vehicle, and does not depend on the environment, namely the inertial data provided by an inertial measurement unit, to overcome the limitations in \textbf{a)} and \textbf{b)}. In order to avoid being limited to one specific type of damage as stated in \textbf{c)}, several types of damage that cover most known cases of failure are considered. Additionally, the proposed solution will be able to detect the damaged propeller, discern the type of damage affecting it, and estimate its magnitude, covering all the cases in \textbf{d)}. Finally, with respect to other works such as \cite{Ghalamchi2020,Baldini2023}, here the information of the commanded torques is also processed to be used as features, showing an improvement in the detection and estimation of the damage, and being crucial for accurate identification of the damaged propeller.

To demonstrate the versatility and effectiveness of the proposed methodology in this work, it will be applied to the dataset introduced in \cite{Baldini2023}. It is noteworthy to emphasize that the dataset features a test scenario with a six-rotor UAV, in contrast to the four-rotor design addressed in our research. This distinction in vehicle configuration underscores the generalization capability of the methodology, showcasing its adaptability and efficacy even across different  configurations. This approach not only reinforces the robustness of the proposed method but also highlights its potential to significantly contribute to damage detection and assessment across a variety of unmanned aerial vehicle platforms.


\subsection{Organization}
This paper is organized as follows: First the problem of interest is stated, followed by a short description of multirotor dynamics. Then, the experimental platform  used to produce and collect flight data, and the nature of the experiments, is described. An analysis of the recorded flight data is then provided, and the architecture to solve the problem is stated. Then, the results are presented, along with other variants that were considered during the process. To provide proof of the ability to generalize the method, the same procedure is applied to an online-available dataset for a vehicle with different topology. Finally, conclusions are drawn, while also presenting some ongoing and future work.

\section{Problem statement}
\begin{figure}[t!]
  \centering
  \includegraphics[width=0.7\linewidth]{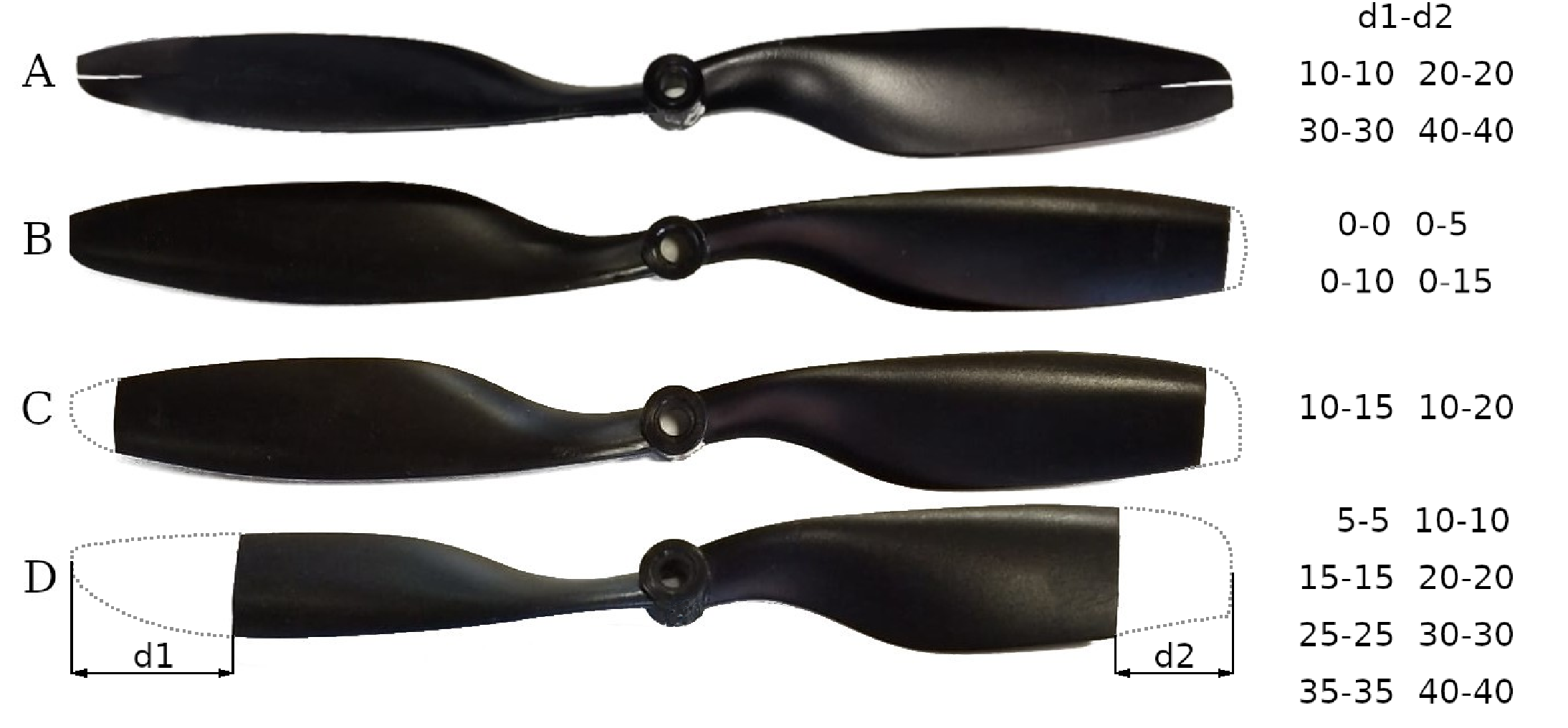}
  \caption{Propellers with, A-longitudinal, B,C-asymmetrical, and D-symmetrical damages. The numbers on the right indicate the cuts performed on each of the 18 propellers used}
  \label{fig:props_all}
\end{figure}

In light of the inherent challenges associated with the vast array of potential damage scenarios that could affect a propeller blade, encompassing dents in all conceivable locations, various deformations, and any type of breakage, practical considerations require a focused approach. The inclusion of every possible damage variant becomes impractical due to resource constraints, including limitations in data collection, computational capacity, and overall project scope. Recognizing these constraints, we have made strategic decisions to enhance the feasibility and relevance of our study. 
By addressing a representative set of scenarios, we aim to provide insights that are both feasible within our resource constraints and relevant to real-world applications. All the possible damage types considered in this work, as well as the magnitude of the damage used for the experimental tests, are shown in Fig. \ref{fig:props_all}.

First, a symmetrical damage is considered, where both tips of the propellers are cut at the same length, as shown in Fig. \ref{fig:props_all}.D . As the geometry of the propeller remains symmetric and balanced, its behavior is similar to that of a healthy propeller, but, as its surface is reduced, so is the air flow if the speed is maintained. From the point of view of the vehicle, it is seen as a loss of thrust and torques that the rotor produces, as it will be described in the following section. This type of failure covers several effects that can appear in a rotor-propeller set, due to wear of the propeller, and wear of the rotor's bearing, which produces friction and also lowers the efficiency.

A second type of damage is one where both tips are also cut, but at different lengths, as shown in Fig. \ref{fig:props_all}.B-C . This type of damage is consistent with the vehicle grazing or softly colliding with another object, where one of the propellers may shortly hit a hard surface and suffer a small breakage in one or both of its tips, which, statistically, will never be equal. While the loss of propeller surface will produce a loss of efficiency just as the first case, the mass imbalance in a high rotating speed propeller will also introduce severe vibrations. This kind of failure could be critical to UAVs, as the vibrations that translate to the structure may render the inertial measurements from the sensors unusable, and in consequence make the estimation of the pose of the vehicle impossible.

The last type of failure analyzed is a longitudinal cut of the same length in both of the tips, as presented in Fig. \ref{fig:props_all}.A This damage could be caused by a collision, or also the separation of fiber carbon layers in propellers made of this material. In this case, no propeller surface is lost, nor is there any kind of mass imbalance. However, the behavior of the propeller at any given speed is unpredictable, as the flapping of the blade may introduce any kind of vibration or lower thrust / torque than expected. This type of failure should be difficult to identify, as its effects may vary depending on the maneuver that is being executed, the wind conditions, and even the temperature that may allow the tips to bend more.

While damage may appear in more than one propeller at the same time, this work will be based on the premise that there is only one damaged propeller, and the objective is to detect where and at which extent.

As the specific characteristics of multirotors vary greatly, particularly regarding number and type of onboard sensors, more useful information for damage detection may be obtained from advanced, complex aircrafts. However, a solution tailored to the capabilities of those vehicles would be severely restrictive for more simple, or even non-commercial vehicles. So, a first restriction for this work is that the information used for detection should come from sensor data or internal variables that are available in any kind of multirotors, regardless its degree of complexity.

An additional restriction is that the proposed method should be able to be extended to other type of multirotor vehicles. While the proposed solution will be trained with a particular set of data collected with a specific vehicle, and thus those results will work only for that particular model, the architecture and procedure can be extended to other vehicles with different number of rotors and geometry.

\section{Vehicle model}
\begin{figure}[t!]
  \centering
  \includegraphics[width=0.65\linewidth]{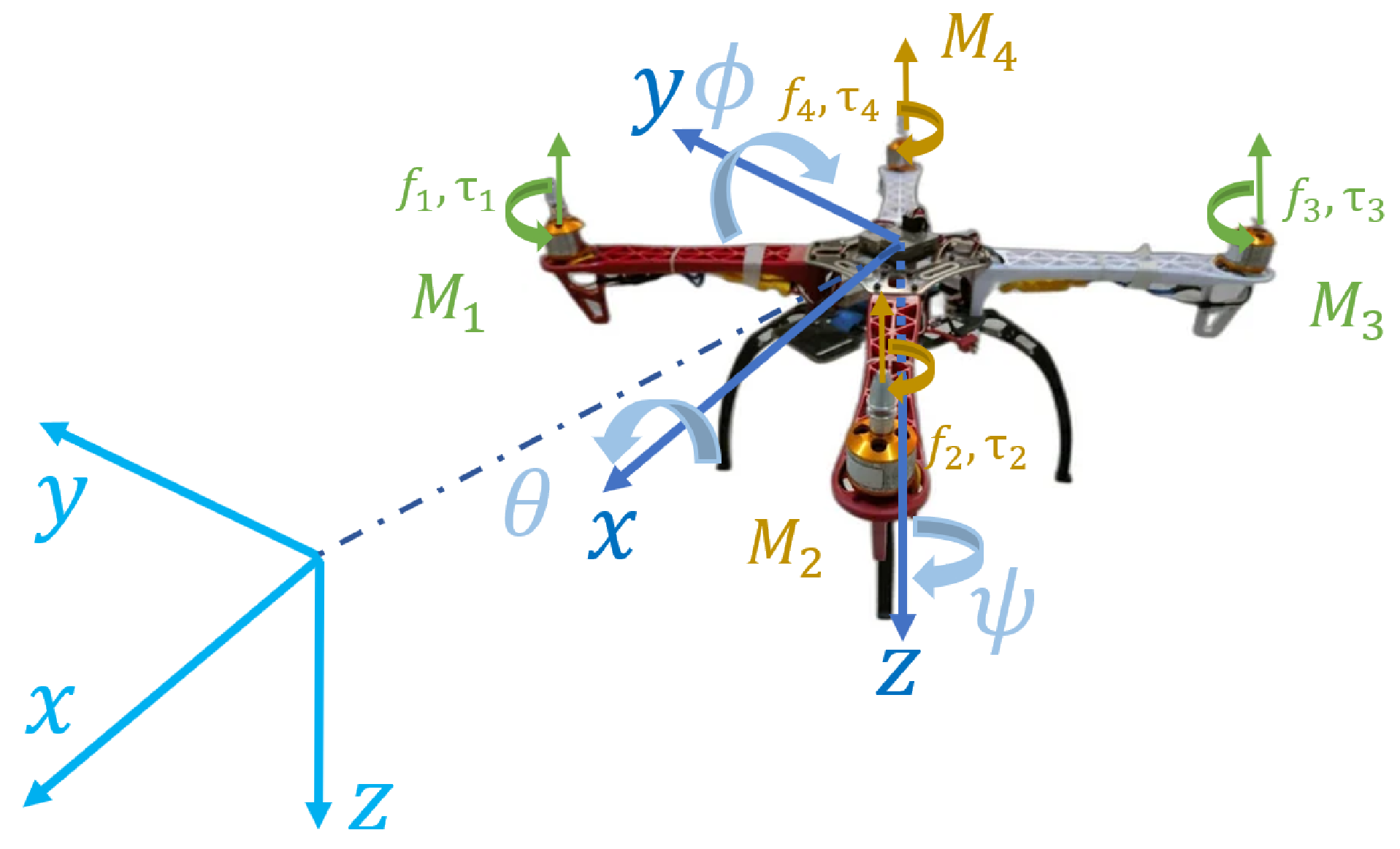}
  \caption{Multirotor with four rotor-propeller sets, with motor numbering, direction of thrust and torque of each actuator, and reference frames.}
  \label{fig:quad_terna}
\end{figure}

Generally, a multirotor is an aircraft with three or more rotors, where the flight control is based on the speed variation of each rotor. In standard commercial vehicles, the structure is commonly composed of a center where the electronic components and power source are mounted, and several arms that extend radially from that center, at which end the rotor-propeller set is located. Then, each rotor is placed at the same fixed distance from the center, and are generally equally spaced in the horizontal plane, as shown in Fig. \ref{fig:quad_terna}. A fixed reference frame is defined in the geometrical center of the vehicle, with the \textbf{x} axis pointing to the nose (front) of the vehicle, the \textbf{y} axis to the right, and the \textbf{z} axis downwards. The rotations along these axis are called roll (represented by the $\theta$ angle), pitch (named the $\phi$ angle), and yaw ($\psi$ angle) respectively.

The actuators are generally arranged such that two neighboring rotors rotate in different directions (clockwise or counterclockwise, looking from above) and with identical but mirrored propellers, so that the thrust is always generated upwards, but the torque produced by conservation of momentum is in opposite directions. This is necessary to produce thrust and independent torque in each of the three axes. The numbering of the rotors is such that rotor 1 is the front-right one, and then the others are numbered in order, counterclockwise.

In a normal state of operation, each unidirectional rotor-propeller set produces a force and torque in the $z$ axis, with a lower limit of zero and an upper boundary given by its maximum rotation speed. 
The joint effect of all rotors' forces and torques generate an upwards vertical thrust, and a set of three torques to rotate the vehicle in any arbitrary direction. The relationship between those sets of magnitudes can be assumed to be linear around the nominal working point, and is given by a matrix \textbf{A} such that $\left[q_x,q_y,q_z,T\right] = A \left[f_1,f_2,..,f_n\right]$, where $q_i$ are the torques in each axis, $T$ is the vertical thrust, and $f_i,\ i\in\left\{0,1,2,...,n\right\}$ are the forces produced by the $n$ rotors. The matrix \textbf{A} is dependent on the geometry of the vehicle (constant if the structure is fixed), namely the location and orientation of the rotors with respect to the center of mass, and the motor constants. For example, for the symmetric quadrotor presented in Fig. \ref{fig:quad_terna}, the matrix $A$ is:

\begin{align*}
    A &=
\begin{bmatrix}
-l cos(\pi/2) & l cos(\pi/2) & l cos(\pi/2) & -l cos(\pi/2) \\
l cos(\pi/2) & l cos(\pi/2) & -l cos(\pi/2) & -l cos(\pi/2) \\
k_t & -k_t & k_t & -k_t \\
-1 & -1 & -1 & -1
\end{bmatrix}
\end{align*}

\noindent where the $k_t$ constant is usually established experimentally.


As special conditions such as propeller damage are considered, the force and torque exerted by the motor may differ with respect to the commanded values. It is this discrepancy which in most cases allows to detect the existence of a failure.
However, most vehicles do not have a way to measure the exact force, torque, or even the actual speed of the rotors, so there are no direct measurements that provide information about the current state of the actuator. Still, a lower exerted force or torque than the one commanded will produce a different maneuver than the one expected, or, similarly, a higher commanded force or torque will be needed to make the same maneuver when a propeller is damaged. This fact will be exploited later to identify which of the rotors is the one presenting a failure.

\section{Data acquisition and selection}\label{sec:dataacq}
\begin{figure}[t!]
  \centering
  \includegraphics[width=0.55\linewidth]{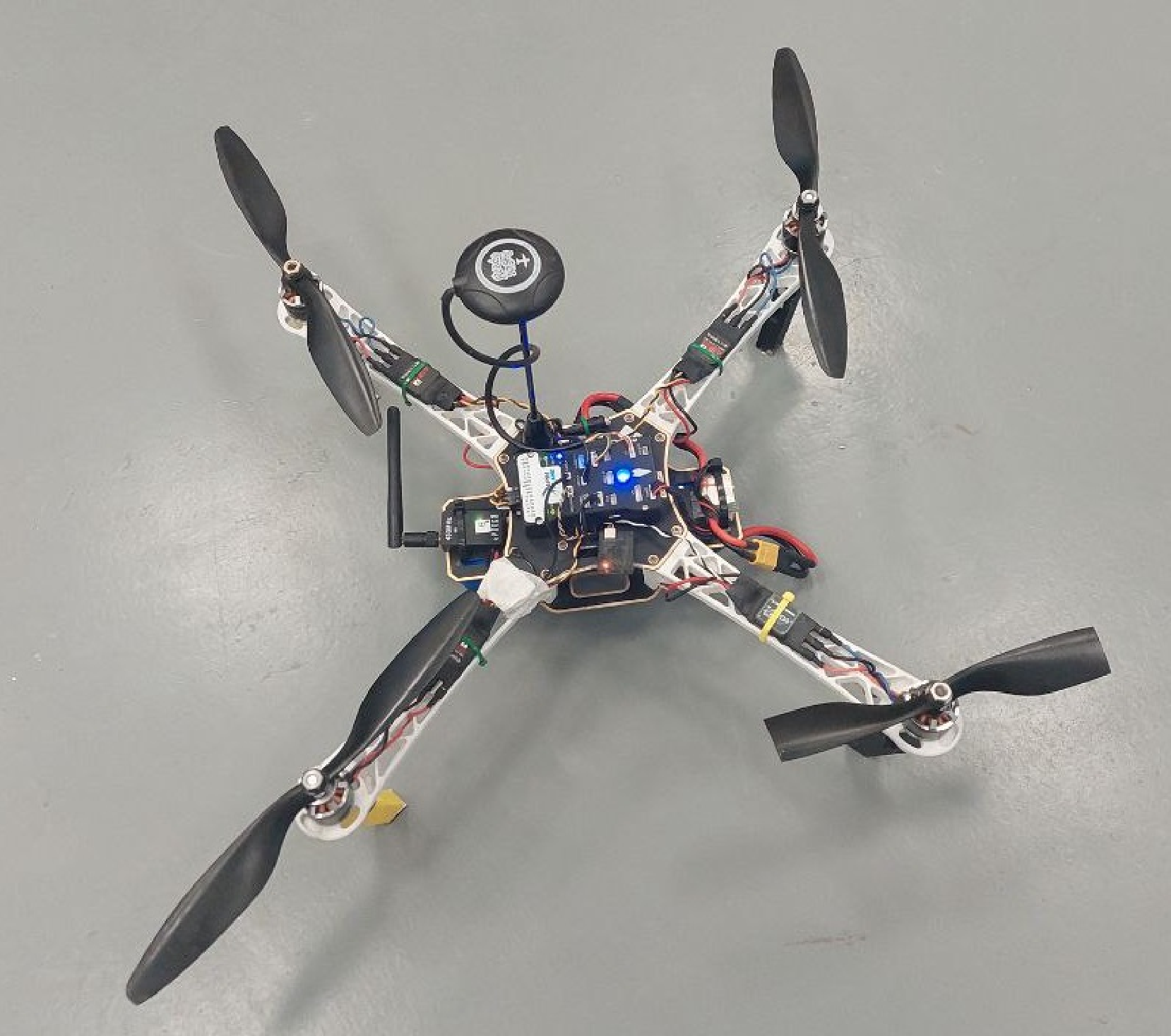}
  \caption{Quadrotor platform used for flight data gathering, with one damaged propeller.}
  \label{fig:quad}
\end{figure}

In order to obtain realistic data, a custom dataset was gathered using a real vehicle, which was assembled as shown in Fig. \ref{fig:quad}.
A quadrotor vehicle was selected as it is the simplest multirotor fixed structure, where all the rotor-propeller pairs are at the same radial distance of the center of the vehicle, equally distributed in the horizontal plane. This particular model is based on the DJI F450 multirotor which is a known commercial frame, with 920KV rotors, and plastic 1045 propellers, capable of \SI{1}{\kilogram} of thrust each. The flight controller is a Pixhawk 2.4.8, with external magnetometer, and a DSMX receiver from Spektrum radio control. The tasks of the flight controller include collecting sensors' measurements, data filtering, attitude estimation, and torque and individual motor control, as well as data logging and receiving commands from an RC controller. The system is powered by a 4s LiPo (\SI{14.8}{\volt}) battery, with the total weight of the system being \SI{1.5}{\kilogram}.

The minimum data needed to fly one of these vehicles is the one provided by an IMU (Inertial Measurement Unit), which provides tridimensional acceleration and gyroscopic measurements. These are needed to estimate the pitch and roll angles of the vehicle in order to stabilize it in the air. While a magnetometer is needed to also estimate the yaw angle, a speed control may be used in that axis based only on the gyroscopic measurements. 

In order to control the attitude of the vehicle, several control algorithms run in the flight controller, which use the inertial measurements to estimate the current attitude, then compare them to the commanded angles in each axis, and generate a set of torques for the vehicle to reach the desired attitude. The set of calculated torques are then transformed into a set of commanded speeds for the rotors, all this information being readily available as different variables inside the flight controller. The last control variable needed to fly the vehicle is the vertical thrust, the combined amount of force that all the rotors exert, and which will define the vertical acceleration. It will generally have an average value equal the weight of the vehicle, which will allow for hovering (vehicle still in the air, not ascending or descending), and will increase or decrease slightly during maneuvers.

The process of gathering a full IMU measurement, calculating the attitude of the vehicle, calculating the torques needed to reach the desired attitude, and commanding the necessary speeds to the rotors is a cycle that is executed periodically at a fixed frequency, which usually varies in the range of \SI{100}{\hertz}-\SI{400}{\hertz} depending on the vehicle. In this particular vehicle, the Pixhawk controller allows to configure a data rate of \SI{222}{\hertz}, logging in its micro SD memory the sensor data and control variables of each cycle. \ch{To replicate the data acquisition process, the user should configure the Pixhawk's SDLOG\_PROFILE 10-bit parameter to value 8, as bit 4 corresponds to the \textbf{High Rate} mode - Full rates for analysis of fast maneuvers (RC, attitude, rates and actuators)}

\ch{With this configuration, several flights were performed in an indoor environment, where in each of them the propeller mounted in motor 1 was changed for each of the damaged propellers listed in Fig. \ref{fig:props_all}}. For each flight with a different damaged propeller, a sequence lasting approximately \SI{120}{\second} was followed, \ch{with experiments using the most damaged propellers being slightly shorter due to increased danger.} Initially, the vehicle took off and remained stationary in the air, only performing slight correction maneuvers for the first \SI{40}{\second}. During the next \SI{40}{\second}, the vehicle executed several soft maneuvers, including slow rotations in each axis and gradual ascents and descents, \ch{with pitch and roll maneuvers of less than \SI{5}{\degree}, and yaw maneuvers of less than \SI{10}{\degree\per\second}}. In the final \SI{40}{\second}, the vehicle underwent aggressive maneuvers in each axis \ch{with combined pitch and roll maneuvers of around \SI{20}{\degree}, and yaw maneuvers of less than \SI{50}{\degree\per\second}}, and abrupt ascents and descents, before finally landing at the starting position. 

All the flights were performed while the damaged propeller was located in rotor number 1, the front-right one. To have a complete dataset for identifying in which rotor the damaged propeller was located, the same set of experiments should have been repeated but changing the location of the damaged propeller.
However, due to the symmetry of the vehicle, it was easy to augment the data and generate the three remaining datasets, one for each motor in failure, by performing an adequate rotation \ch{of the accelerometer, gyroscope and torque data} around the $z$ axis of the vehicle.

The recorded flight data as well as the algorithms used to train and test the proposed solution can be found at \cite{QuadCarbDataset}.

\section{Propeller damage and effects}
\begin{figure*}[t!]
  \centering
  \includegraphics[width=0.98\textwidth]{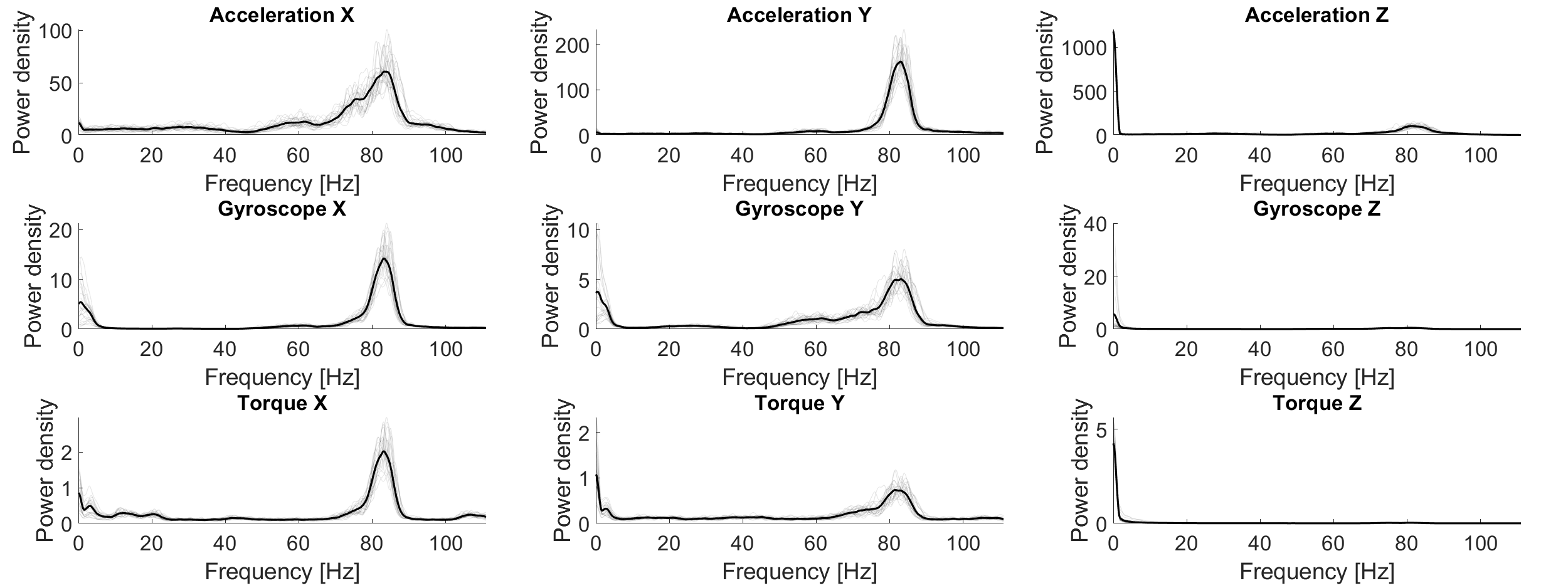}
  \caption{Average power density over the frequency spectrum for the inertial sensors and the torque control signals for a flight with four healthy propellers (dashed). In transparency, the power density taken in several smaller sections of the same flight. The power density is almost the same, both if the vehicle is hovering, performing a maneuver, or when considering the full flight.}
  \label{fig:specgram_nofail}
\end{figure*}

As it was described in the problem statement, different types and degrees of damage in a propeller produce different effects over the aircraft, which may be captured by different sensors or introduced in the calculations of the control variables. A perfectly healthy vehicle, flying still in the air without maneuvers, would have a thrust force equal to the weight of the vehicles, and zero torque on all axis as it is not rotating. This means that all rotors are spinning at the same speed (exerting he same force and torque), and there would appear vibrations in the structure related to the speed of the rotors. For example, for this vehicle that weights \SI{1.5}{\kilogram}, each rotor should produce a thrust of \SI{0.375}{\kilogram}, as its maximum thrust is \SI{1}{\kilogram}, it should be spinning at around 0.375 of its maximum speed to produce such thrust. Its maximum speed is 920KV times the battery voltage \SI{14.8}{\volt}, which gives 13616rpm, and 0.375 of that value is 5106rpm or \SI{85}{\hertz}. Fig. \ref{fig:specgram_nofail} shows in dashed black line the average power density over the frequency spectrum for the full flight using four healthy propellers, for each of the accelerometers, gyroscopes, and control torques, where the energy at each frequency is the average recorded from the take-off till landing. It can be seen that there is a distinctive peak at around \SI{83}{\hertz}, which corresponds to the effect described above. While there is also energy in the low frequency bands, it does not correspond to vibrations but rather to the maneuvers (both soft and aggressive) performed during flight. Note that while the magnitude of the energies for all variables are similar in the \textbf{x} and \textbf{y} axes, the energy in the \textbf{z} axis is lower around the \SI{83}{\hertz} band and higher at lower frequencies. This is because, as the moment of inertia is higher in the vertical axis, and the required variation in the speed of the rotors is much higher to execute any maneuver, the effect of the vibrations introduced is diminished while the maneuvers require more energy. Also, the accelerometer in that axis constantly measures the force of gravity, introducing a great energy in the zero frequency. The transparent lines in the same figure show the power density taken in several smaller sections (\SI{6}{\second} window) of the same flight, to show that the frequency spectrum is similar during hovering, and during soft and aggressive maneuvering.

\begin{figure*}[t!]
  \centering
  \includegraphics[width=0.98\textwidth]{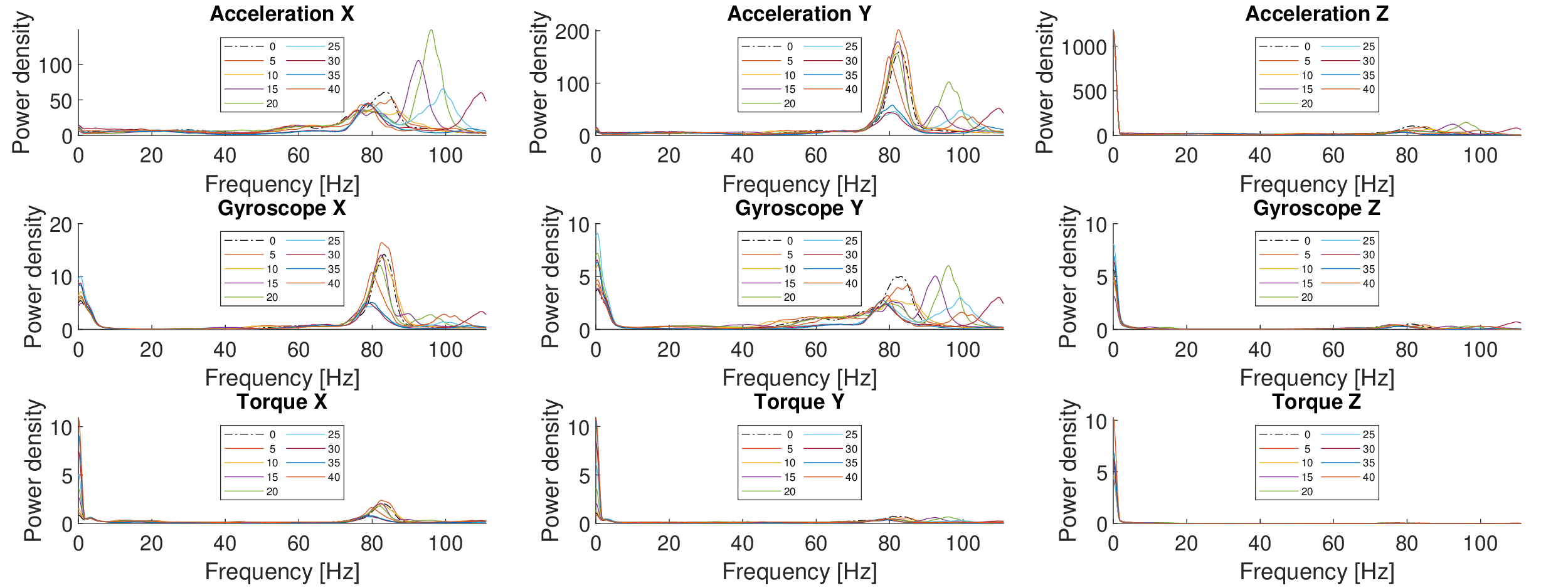}
  \caption{Frequency spectrum for the inertial sensors and the torque control signals for a flight with a symmetrically cut propeller in rotor 1. Healthy flight as reference in dashed black line.}
  \label{fig:specgram_symm}
\end{figure*}

In the case of a symmetric damage to a propeller, a loss of thrust would show up if the rotor maintains its speed, which would be equivalent to an external force exerted downwards on that rotor. As a consequence, the vehicle would seem to be subjected to a perturbation torque in some direction (which would depend on the location of the damaged propeller). 
The flight controller must command a constant torque to counteract this perturbation and keep the vehicle stable in the air. Consequently, a constant bias becomes evident in the torque control signals. In order for the damaged propeller to reach again the original thrust it was producing, its rotor should increase its speed, and due to the new propeller shape it may introduce vibrations due to the interaction with the air vortex generated by the tips. Then, effects such as the ones presented in Fig. \ref{fig:specgram_symm} are expected, which shows the frequency spectrum for flights where both tips of the propeller in rotor 1 are cut by a length between \SI{5}{\milli\meter} and \SI{40}{\milli\meter} in steps of \SI{5}{\milli\meter}. 

Note that the energy peaks for the propellers cut more than \SI{30}{\milli\meter} fall outside the observable spectrum (\SI{111}{\hertz} for a sampling rate of \SI{222}{\hertz}), as the rotor spinning speed exceeds the nyquist frequency of the sampled data. However, notice also that secondary peaks emerge in each figure at a lower frequency that those of the healthy case, as some of the non-damaged propellers lower their spinning speed to compensate the torque missing from the damaged one. The results of this work will show that the secondary peaks are also good to detect and identify the occurrence of damage on a propeller, but it is worth mentioning that increasing the sampling rate could provide richer and more precise information for damage identification (for example, the Pixhawk flight controller has a high logging data rate of \SI{666}{\hertz}, but not for all data). However, if considering the implementation of the algorithm in an onboard computer, increasing the data rate also increases the computational load and processing time to detect and identify quickly the occurrence of a failure, so it is of interest to try and reach a solution without greatly increasing the sampling rate.

\begin{figure*}[t!]
  \centering
  \includegraphics[width=0.98\textwidth]{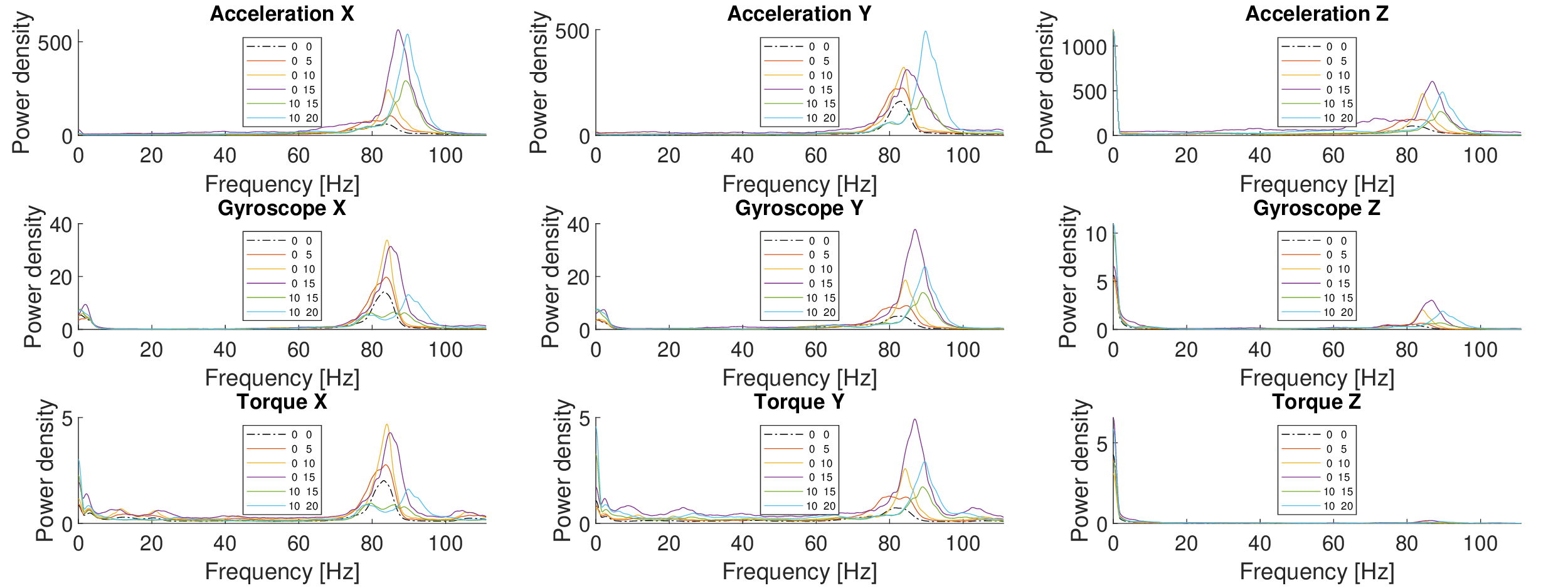}
  \caption{Frequency spectrum for the inertial sensors and the torque control signals for a flight with a asymmetrically cut propeller in rotor 1. Healthy flight as reference in dashed black line.}
  \label{fig:specgram_asymm}
\end{figure*}

For the asymmetrically cut propellers, the same loss of thrust is expected as the surface is still being reduced, so the damaged propeller will increase its speed to compensate this loss, but it will also introduce severe vibrations. A healthy propeller is usually perfectly balanced, either from factory, or by the mechanic / pilot, through a meticulous process, similar to that used for balancing the wheels of a car. The reason is that mass imbalance at high speeds produces vibrations that may be destructive to the structure, and, in the case of UAVs, may introduce vibrations in the sensors to the point of rendering them unusable. This fact will be used to detect and identify this type of failure. The frequency spectrum for a series of flights with one asymmetrically cut propeller is shown in Fig. \ref{fig:specgram_asymm}, where the labels indicate the millimeters cut from each tip of the same propeller (for example, the green line is for a propeller cut \SI{10}{\milli\meter} in one tip and \SI{15}{\milli\meter} in the other). Note that, as the sum of both cuts increase, the surface of the propeller is reduced, and the frequency of the energy peak increases as the damaged rotor speed does the same. Also, as the difference between both cuts increases, so does the mass imbalance and the vibrations that this produces, so the energy at the peak has a higher value. 
In the flights where one propeller tip was cut by \SI{15}{\milli\meter} and the other was not cut, the resulting high vibrations caused the inertial measurements to become extremely noisy. This led to imprecise attitude estimations and resulted in several crashes during maneuvers.

\begin{figure*}[t!]
  \centering
  \includegraphics[width=0.98\textwidth]{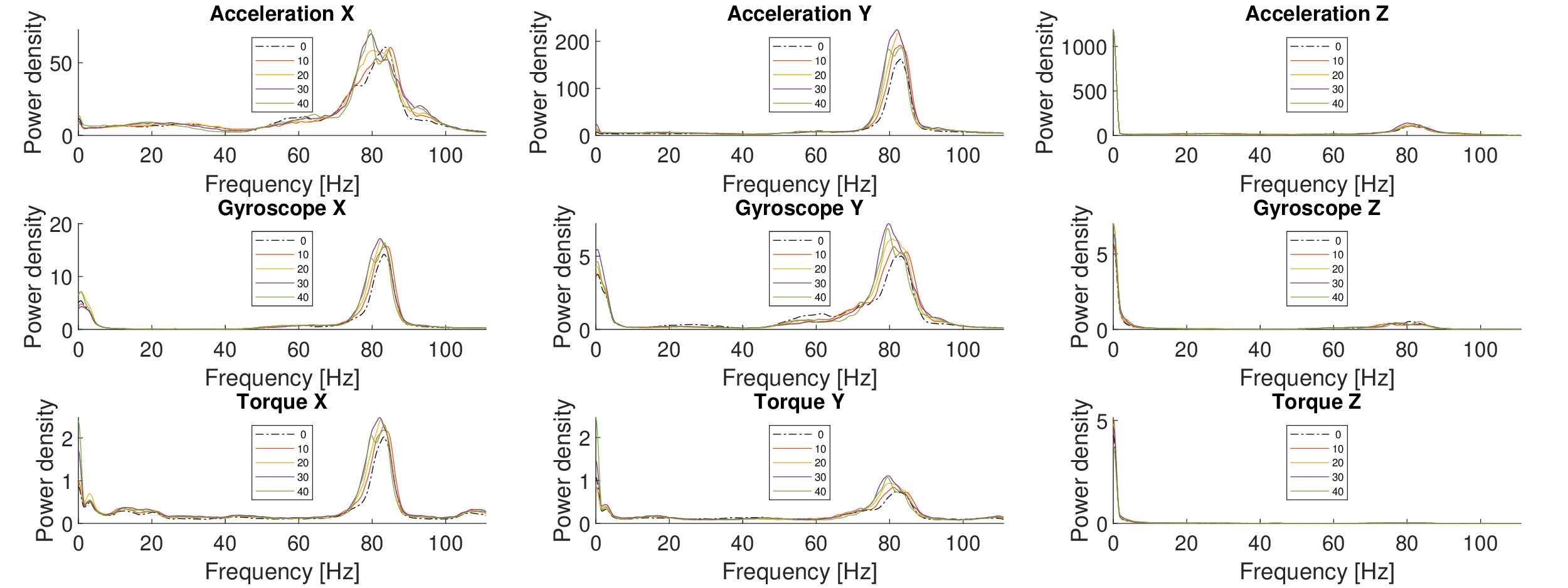}
  \caption{Frequency spectrum for the inertial sensors and the torque control signals for a flight with a longitudinal cut propeller in rotor 1. Healthy flight as reference in dashed black line.}
  \label{fig:specgram_long}
\end{figure*}

Lastly, it was previously unknown what effect regarding thrust and vibrations would be produced by a longitudinal cut on a propeller, so the following will be conjectures based on the recorded data. As seen in Fig. \ref{fig:specgram_long}, which presents the frequency spectrums for the flights with such damage considering cuts \SI{10}{\milli\meter} to \SI{40}{\milli\meter} deep in both tips (in steps of \SI{10}{\milli\meter}), the frequency of the peak decreases slightly, while the value of the peak increases. As there is not mass imbalance, vibrations similar to those of the asymmetric damage were not expected, but is still remarkable that almost no additional vibration is introduced, considering the magnitude of the damage. On the other hand, it is difficult to know exactly the reason for the decrease in the peak frequency, as it should be correlated with the speed of the damaged propeller. Then, it is possible that the damaged propeller is indeed reducing its speed to produce the same thrust as a healthy propeller. This may be caused by a number of different effects: one is that the surfaces of the propeller or the aerodynamic boundary layers are separating at high speeds, virtually increasing the surface and the thrust produced; something similar would happen if, due to the damage, the blade is twisting in such a way that it increases its pitch angle, also increasing the thrust produced.

Considering the nature of the perturbations introduced by the different types of damage, it is reasonable to work with the data in the frequency domain, as it is easier to spot the occurrence of vibrations at different frequencies. In order to avoid an infinite number of points that represent the frequency spectrum of each sensor or command signal, a finite number of equally spaced bands is defined over the spectrum, and the energy of each band will be used as feature for damage detection and identification. In order to produce a frequency spectrum, the Fourier transform over a time window must be performed, so a window was selected consisting of 222 cycles of consecutive inertial and control data, corresponding to \SI{1}{\second} of recorded information, which will be referred as a sample from this point. Then, the features extracted from the frequency spectrum will correspond to the energy in each band during that time window. Also, as a propeller sees its surface modified, a constant compensation torque is generated in some direction, and the speed variation of the rotor required for different maneuvers may severely change with respect to the healthy case. Then, the mean, variance, skewness and Kurtosis values for each of the three commanded torques will be added to the feature set.

\st{In order to avoid obtaining samples that provide identical features, an overlap of 32 timesteps is allowed, so the samples are sufficiently separated over time.}
\ch{Each sample is taken with a difference of 32 timesteps, equivalent to around \SI{144}{\milli\second} (works such as \cite{Etchemaite2023} show that even a \SI{200}{\milli\second} delay in the detection of a failure can be enough to take corrective actions). This allows the samples to be sufficiently separated over time, but also considers that the detection is performed at a high enough frequency to monitor the state of the vehicle during flight. Table \ref{tab:number_samples} summarizes the number of samples obtained from each experiment with a different damaged propeller located always in motor number 1, where it can be noted that flights with higher degree of damage have a lower number of samples due to the difficulties in performing a long, safe data acquisition process. Additionally, after performing data augmentation to generate synthetic flight data for the damaged propellers located in the other motors, the final number of samples for each type of propeller is increased four times.}

\renewcommand{\arraystretch}{1.2}
\begin{table}[]
\resizebox{\columnwidth}{!}{ 
\begin{tabular}{ccccccccccc}
\hline
\multicolumn{11}{|c|}{Propeller type and cuts {[}mm-mm{]} / Number of samples [\#]}\\ \hline
\multicolumn{1}{|c|}{Healthy} & \multicolumn{1}{c|}{\#}   & \multicolumn{1}{c|}{} & \multicolumn{1}{c|}{Symm}  & \multicolumn{1}{c|}{\#}   & \multicolumn{1}{c|}{} & \multicolumn{1}{c|}{Asymm} & \multicolumn{1}{c|}{\#}  & \multicolumn{1}{c|}{} & \multicolumn{1}{c|}{Long}  & \multicolumn{1}{c|}{\#}  \\ \cline{1-2} \cline{4-5} \cline{7-8} \cline{10-11} 
\multicolumn{1}{|c|}{0-0}     & \multicolumn{1}{c|}{1005} & \multicolumn{1}{c|}{} & \multicolumn{1}{c|}{5-5}   & \multicolumn{1}{c|}{1098} & \multicolumn{1}{c|}{} & \multicolumn{1}{c|}{0-5}   & \multicolumn{1}{c|}{847} & \multicolumn{1}{c|}{} & \multicolumn{1}{c|}{10-10} & \multicolumn{1}{c|}{863} \\ \cline{1-2} \cline{4-5} \cline{7-8} \cline{10-11} 
                              &                           & \multicolumn{1}{c|}{} & \multicolumn{1}{c|}{10-10} & \multicolumn{1}{c|}{1043} & \multicolumn{1}{c|}{} & \multicolumn{1}{c|}{0-10}  & \multicolumn{1}{c|}{879} & \multicolumn{1}{c|}{} & \multicolumn{1}{c|}{20-20} & \multicolumn{1}{c|}{954} \\ \cline{4-5} \cline{7-8} \cline{10-11} 
                              &                           & \multicolumn{1}{c|}{} & \multicolumn{1}{c|}{15-15} & \multicolumn{1}{c|}{1131} & \multicolumn{1}{c|}{} & \multicolumn{1}{c|}{0-15}  & \multicolumn{1}{c|}{465} & \multicolumn{1}{c|}{} & \multicolumn{1}{c|}{30-30} & \multicolumn{1}{c|}{875} \\ \cline{4-5} \cline{7-8} \cline{10-11} 
                              &                           & \multicolumn{1}{c|}{} & \multicolumn{1}{c|}{20-20} & \multicolumn{1}{c|}{1051} & \multicolumn{1}{c|}{} & \multicolumn{1}{c|}{10-15} & \multicolumn{1}{c|}{885} & \multicolumn{1}{c|}{} & \multicolumn{1}{c|}{40-40} & \multicolumn{1}{c|}{968} \\ \cline{4-5} \cline{7-8} \cline{10-11} 
                              &                           & \multicolumn{1}{c|}{} & \multicolumn{1}{c|}{25-25} & \multicolumn{1}{c|}{902}  & \multicolumn{1}{c|}{} & \multicolumn{1}{c|}{10-20} & \multicolumn{1}{c|}{766} &                       &                            &                          \\ \cline{4-5} \cline{7-8}
                              &                           & \multicolumn{1}{c|}{} & \multicolumn{1}{c|}{30-30} & \multicolumn{1}{c|}{784}  &                       &                            &                          &                       &                            &                          \\ \cline{4-5}
                              &                           & \multicolumn{1}{c|}{} & \multicolumn{1}{c|}{35-35} & \multicolumn{1}{c|}{763}  &                       &                            &                          &                       &                            &                          \\ \cline{4-5}
                              &                           & \multicolumn{1}{c|}{} & \multicolumn{1}{c|}{40-40} & \multicolumn{1}{c|}{752}  &                       &                            &                          &                       &                            &                          \\ \cline{4-5}
\multicolumn{1}{l}{}          & \multicolumn{1}{l}{}      & \multicolumn{1}{l}{}  & \multicolumn{1}{l}{}       & \multicolumn{1}{l}{}      & \multicolumn{1}{l}{}  & \multicolumn{1}{l}{}       & \multicolumn{1}{l}{}     & \multicolumn{1}{l}{}  & \multicolumn{1}{l}{}       & \multicolumn{1}{l}{}    
\end{tabular}
}
\caption{Number of samples (\SI{1}{\second} windows) for each flight with healthy or damaged propellers.}
\label{tab:number_samples}
\end{table}
\renewcommand{\arraystretch}{1}

\section{Proposed architecture}
\ch{Initially, a single neural network and classifier were proposed to identify the type of failure and the degree of damage, but proved difficult to separate the longitudinal cuts from the other two types. The neural network approach worked well if each type of damage was treated separately (training a different one for each), and several tests showed that also there was no noticeable difference in the inference when treating symmetric and asymmetric damage together. This limitation also applied to training classifiers to locate the affected propeller, and worked best when treating the longitudinal damage separately from the other types. Then, one neural network and classifier was needed for symmetric and asymmetric damage, and another ones for longitudinal damage. This also implied that another classifier was required to first detect the type of damage, and only then use the corresponding neural network/classifier pair. While the solution requires more memory and processing power, it also allows to easily extend the problem to new types of damage, as only the damage type classifier needs to be re-trained when adding new types of failure (this would not apply if the intention is to classify the new type of damage together with an existing one, as it is done with the symmetric and asymmetric damage).} 

\begin{figure*}[t!]
  \centering
  \includegraphics[width=0.8\textwidth]{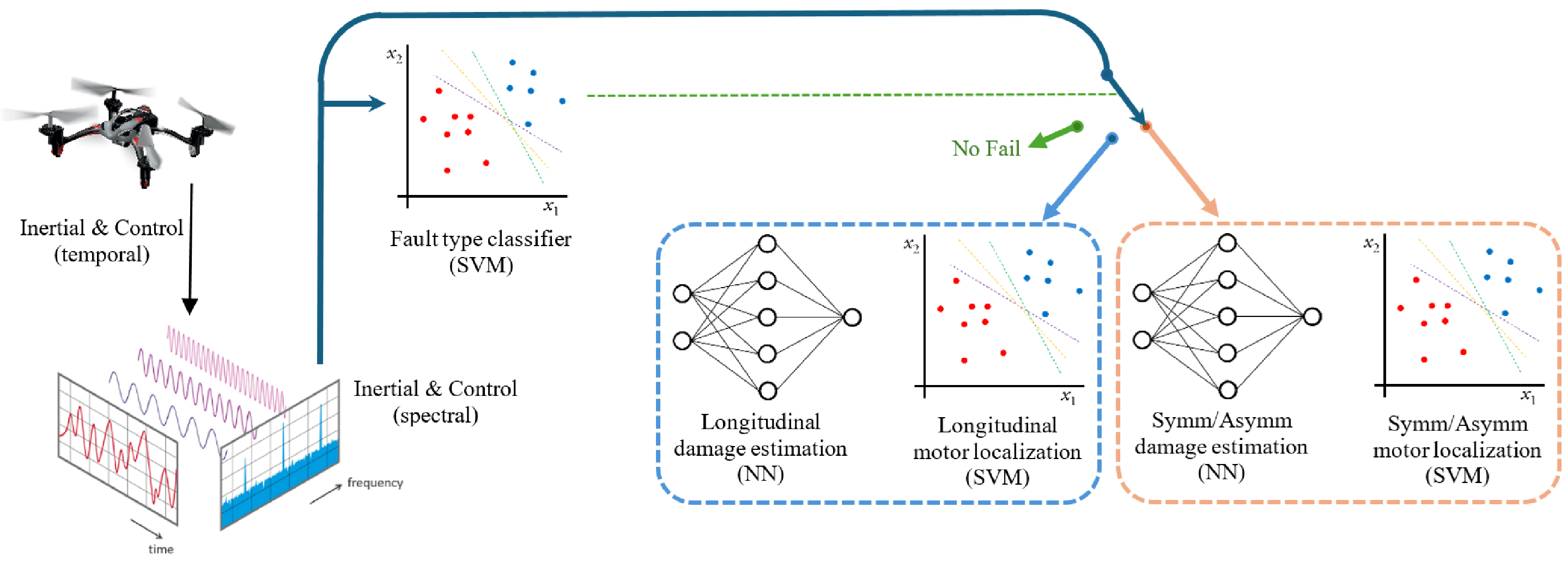}
  \caption{Proposed architecture for propeller damage detection, classification and estimation.}
  \label{fig:architecture}
\end{figure*}

After several modifications to an initially proposed structure, the details of which are summarized in Section \ref{sec:variations}, an architecture as the one shown in Fig. \ref{fig:architecture} was deemed the most adequate.

The features used for all the processes are the same, consisting in the power of the energy bands for each sensor and the mean, variance, skewness and Kurtosis values of the three commanded torques. The width of the energy bands in which the energy is calculated was selected as \SI{5}{\hertz}, for a total of 22 bands in the \SI{111}{\hertz} spectrum. The total number of features is equal to the number of energy bands taken from the spectrum, times 10 (three accelerometers, gyroscopes, and torques, and the vertical thrust), plus the mean, variance, skewness and Kurtosis values for the three torques. This gives a total of 232 features extracted from each sample.

\ch{The dataset was split randomly in 40\%-30\%-30\% of samples for the training, validation, and test dataset}.
To process the features, first a classifier detects if there exists or not a damaged propeller, and if it is a symmetric / asymmetric damage, or a longitudinal one. A support vector machine approach was used, as the monotonous increase in the spectrum peak energy and frequency with the degree of the damage hinted that it could be a linear separable problem. Indeed, the linear kernel worked best than others when testing different possibilities. In this particular classifier, as a single flight with healthy propellers was made, while four with longitudinal damage and 13 with symmetrical and asymmetrical damage were performed, there is a great imbalance in data, as each flight roughly contributes the same amount of samples. \ch{Then, a K-means clustering procedure was performed, reducing the dataset to a group of clusters with the same amount of representative samples. As the healthy propeller flight was the one with the lowest number of samples with a total of 1608 (1005 real samples times 4, times 0.4), a K-means procedure was performed both for the symmetric/asymmetric and for the longitudinal training data to generate 1608 centroids for each, which will be used as synthetic samples. While not very probable as the samples are \SI{1}{\second} wide, there might appear a sample with spectral properties quite different from the median, due to some particular maneuver or sustained perturbation. To avoid having these samples  decrease the performance, the K-means method was used instead of undersampling.}

Then, for each possible class of damage, a neural network was trained to estimate its magnitude, and a classifier to detect in which on the four rotors the damaged propeller was. For the symmetric / asymmetric damage, the neural network has two outputs, one is the sum of the length cut of both tips, and the other is the absolute difference between them (both in millimeters, positive), to have a representative measure of the surface loss and imbalance in a propeller. In the longitudinal case, there is a single output that is the sum of the cuts in both tips, in order to have a similar approach to the former case. 

In the case of detecting in which motor the damage is located, there is now no need to balance the classes, as the flights with a failure in motor 1 were used to produce through data augmentation the flights with a failure in the other motors, so all classes have exactly the same amount of samples. \ch{Still, as the number of samples was large (11352 for each of the four classes) and training took a long time, another K-means clustering was performed to generate synthetic samples, reducing the number of points to 4000 for each class, and speeding up the training process}. The performance difference between a SVM classifier trained with the full dataset and another trained with the reduced one was negligible, and allowed to quickly explore different configurations.

\ch{The neural networks were solved using the \textit{torch} library. Their structures were tested using 1 to 4 hidden layers, with a number of neurons using rule of thumb, including the two-thirds rule and halving. Learning rates were tested using grid search with values ranging from 0.01 to 0.05 in steps of 0.01, and from 0.05 to 0.5 in steps of 0.05. Several optimizers were tested, including Adadelta, Adam, Adagrad and SGD.}

The best solution found was using an architecture consisting of 232 inputs, and three hidden layers with 32, 8 and 4 neurons with ReLU activation function. To train them, the MSE loss function was used, as well as the Adadelta optimizer, with a learning rate of 0.1 to obtain a consistently decreasing, smooth evolution of the loss over 200 epochs. 

The SVM classifiers were trained using the \textit{sklearn} library for \textit{python}, particularly the \textit{sklearn.svm.SVC} pipeline. \ch{Linear, second to fourth order polynomials, RBF and sigmoid kernels were tested, with tolerance ranging from 1e-3 to 1e-5. Eventually, the best solution found used a linear kernel and added the probability output for better analysis.}

\section{Training and results}
This section presents the results obtained for the different blocks of the proposed architecture, considering the particular structure and training methods described in the former section.

\subsection{Damage type classification}
In Table \ref{tab:classif_damagetype} the confusion matrix for the damage classifier is presented, separated by flight. Each row corresponds to a single flight with a different damaged propeller. The \textit{C0} class corresponds to a flight with no damaged propellers, the \textit{C1} class to a symmetric or asymmetric damaged propeller, and \textit{C2} to a propeller with longitudinal damage.



\begin{table}[]
\centering
\begin{tabular}{|cc|ccc|}
\hline
\multicolumn{2}{|c|}{Percentage [\%]}                             & \multicolumn{3}{c|}{Predicted class}                                                          \\ \hline
\multicolumn{1}{|c|}{True Class}           & Damage & \multicolumn{1}{c|}{C0}              & \multicolumn{1}{c|}{C1}              & C2              \\ \hline
\multicolumn{1}{|c|}{C0}                   & 0-0    & \multicolumn{1}{c|}{\textbf{97.71}} & \multicolumn{1}{c|}{0.89}          & 1.39          \\ \hline
\multicolumn{1}{|c|}{\multirow{13}{*}{C1}} & 5-5    & \multicolumn{1}{c|}{34.85}          & \multicolumn{1}{c|}{\textbf{62.51}} & 2.64          \\ \cline{2-5} 
\multicolumn{1}{|c|}{}                     & 10-10  & \multicolumn{1}{c|}{10.44}          & \multicolumn{1}{c|}{\textbf{86.97}} & 2.59          \\ \cline{2-5} 
\multicolumn{1}{|c|}{}                     & 15-15  & \multicolumn{1}{c|}{0.88}          & \multicolumn{1}{c|}{\textbf{98.50}} & 0.62          \\ \cline{2-5} 
\multicolumn{1}{|c|}{}                     & 20-20  & \multicolumn{1}{c|}{0.00}          & \multicolumn{1}{c|}{\textbf{99.62}} & 0.38          \\ \cline{2-5} 
\multicolumn{1}{|c|}{}                     & 25-25  & \multicolumn{1}{c|}{0.22}          & \multicolumn{1}{c|}{\textbf{99.34}} & 0.44          \\ \cline{2-5} 
\multicolumn{1}{|c|}{}                     & 30-30  & \multicolumn{1}{c|}{1.40}          & \multicolumn{1}{c|}{\textbf{96.69}} & 1.91          \\ \cline{2-5} 
\multicolumn{1}{|c|}{}                     & 35-35  & \multicolumn{1}{c|}{0.13}          & \multicolumn{1}{c|}{\textbf{92.93}} & 6.94          \\ \cline{2-5} 
\multicolumn{1}{|c|}{}                     & 40-40  & \multicolumn{1}{c|}{0.27}          & \multicolumn{1}{c|}{\textbf{99.47}} & 0.27          \\ \cline{2-5} 
\multicolumn{1}{|c|}{}                     & 0-5    & \multicolumn{1}{c|}{6.49}          & \multicolumn{1}{c|}{\textbf{42.92}} & 50.59          \\ \cline{2-5} 
\multicolumn{1}{|c|}{}                     & 0-10   & \multicolumn{1}{c|}{4.43}          & \multicolumn{1}{c|}{\textbf{93.98}} & 1.59          \\ \cline{2-5} 
\multicolumn{1}{|c|}{}                     & 0-15   & \multicolumn{1}{c|}{0.00}          & \multicolumn{1}{c|}{\textbf{98.28}} & 1.72          \\ \cline{2-5} 
\multicolumn{1}{|c|}{}                     & 10-15  & \multicolumn{1}{c|}{8.24}          & \multicolumn{1}{c|}{\textbf{81.26}} & 10.50          \\ \cline{2-5} 
\multicolumn{1}{|c|}{}                     & 10-20  & \multicolumn{1}{c|}{0.26}          & \multicolumn{1}{c|}{\textbf{99.74}} & 0.00          \\ \hline
\multicolumn{1}{|c|}{\multirow{4}{*}{C2}}  & 10-10  & \multicolumn{1}{c|}{4.40}          & \multicolumn{1}{c|}{2.88}          & \textbf{93.06} \\ \cline{2-5} 
\multicolumn{1}{|c|}{}                     & 20-20  & \multicolumn{1}{c|}{2.72}          & \multicolumn{1}{c|}{1.99}          & \textbf{95.29} \\ \cline{2-5} 
\multicolumn{1}{|c|}{}                     & 30-30  & \multicolumn{1}{c|}{2.63}          & \multicolumn{1}{c|}{4.00}          & \textbf{93.38} \\ \cline{2-5} 
\multicolumn{1}{|c|}{}                     & 40-40  & \multicolumn{1}{c|}{2.06}          & \multicolumn{1}{c|}{1.96}          & \textbf{95.98} \\ \hline
\end{tabular}
\caption{Performance in damage estimation for symmetric and asymmetric damage.}
\label{tab:classif_damagetype}
\end{table}

The results show a very good performance classifying the samples corresponding to a flight with healthy propellers, as there is a high probability for that class for all the samples taken during that flight.

For the case of symmetric damage, during most of the flights, particularly those corresponding to a high degree of damage, the performance is also very good, with a very high probability of belonging to the correct class. However, for the flight with \SI{10}{\milli\meter} cuts in both tips of a propeller, the performance is lower during some parts of the flight (particularly at the beginning when the vehicle is not performing maneuvers), where it shows a higher probability of belonging to the no damage class. The performance decreases even more in the flight with \SI{5}{\milli\meter} cuts, where only 62.5\% of the samples is correctly classified, and 34.8\% are classified as no failure. Still, this is expected, as it is very difficult to discern between a perfectly healthy propeller, and one with a minimal loss of efficiency which is not producing noticeable vibrations. To put it in an adequate context, the \SI{5}{\milli\meter} propeller should behave almost the same that the healthy propeller, but with around a 3\% efficiency loss in thrust and torque.

For the asymmetric damage flights, again the performance of the classifier is better as the degree of the damage increases, which is logical as the vibrations affect more the sensors and are easier to detect. Only for the case with the lowest damage (\SI{0}{\milli\meter} and \SI{5}{\milli\meter} cuts in a propeller) the samples of that flight may be erroneously classified as a longitudinal damage. This may be due to the vibrations being low in energy, as the mass imbalance is low, and the rotor speed doesn't change much as the efficiency remains almost the same. It would be also expected to erroneously classify these samples as a no failure case.

Lastly, the performance is good for the samples of the longitudinal damage flights.

The results presented in the table correspond to flights with the damaged propeller located in rotor 1, as the results for all the other rotors are similar.


\subsection{Symmetric and asymmetric damage estimation and localization}
\begin{figure*}[t!]
  \centering
  \includegraphics[width=0.98\textwidth]{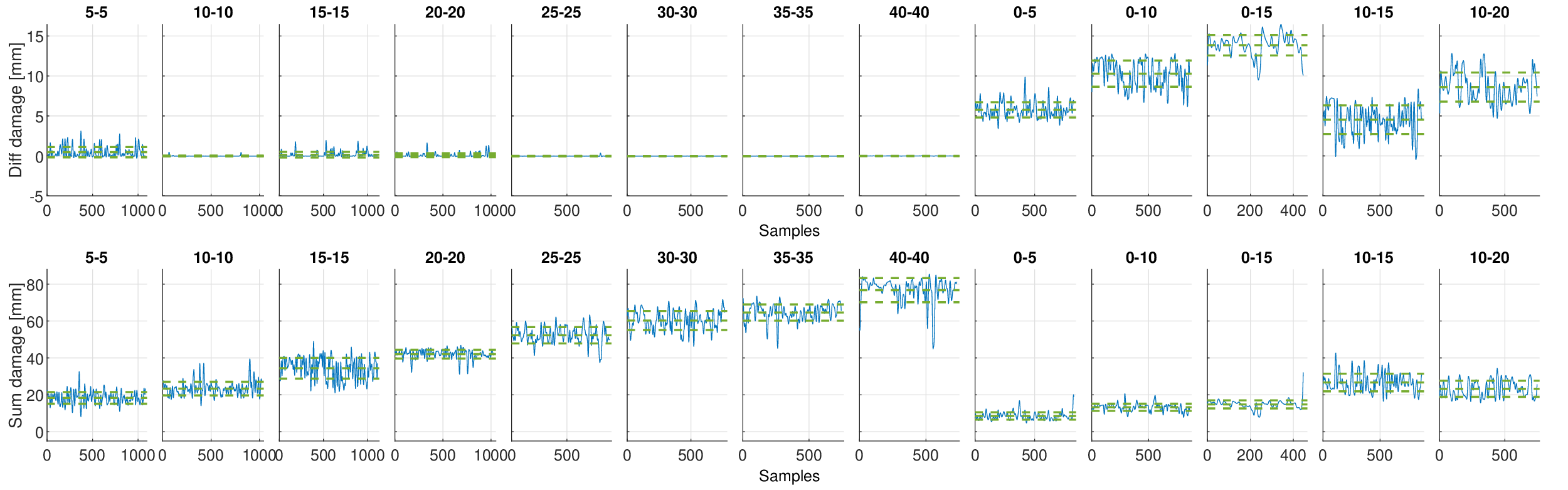}
  \caption{Estimation of the sum and difference in damage for the symmetric and asymmetric damage flights.}
  \label{fig:results_nn_symmasymm}
\end{figure*}

If a sample is correctly classified as corresponding to a symmetric or asymmetric damage, then the same features are used to estimate the magnitude and location of the damaged propeller. \ch{An example of how the network would perform during a full flight is presented in Fig. \ref{fig:results_nn_symmasymm}. The first row corresponds to the predicted difference in damage between the tips for each sample, and the second row to the sum of the damage. Each column corresponds to a full flight with one of the damaged propellers, the first 8 corresponding to flights with symmetric damage, and the last 5 to the ones with asymmetric damage. The titles of each column represent the length cut in both tips of the propeller (i.e. the column titled \textbf{[10,15]} corresponds to a full flight with one damaged propeller which tips were cut \SI{10}{\milli\meter} and \SI{15}{\milli\meter} each.). The horizontal green dashed lines show the mean and plus/minus variance of the prediction for each flight. It should be taken into account that samples used for training, validation and testing are mixed in this figure, so it should not be taken as a metric of performance but rather a demonstration of how it would work in a continuous flight.}


\begin{table}[t!]
\center
\begin{tabular}{|c|cc|cc|cc|}
\hline
\multicolumn{1}{|c|}{\multirow{2}{*}{Cuts}}  & \multicolumn{2}{c|}{NN outputs} & \multicolumn{2}{c|}{Error}             & \multicolumn{2}{c|}{Std. Dev.}       \\ \cline{2-7} 
\multicolumn{1}{|c|}{}  & \multicolumn{1}{c|}{Sum} & Diff  & \multicolumn{1}{c|}{Sum} & Diff    & \multicolumn{1}{c|}{Sum}    & Diff   \\ \hline
5-5   & \multicolumn{1}{c|}{18.2394}    & 0.4586    & \multicolumn{1}{c|}{8.2394}  & 0.4586  & \multicolumn{1}{c|}{3.4454} & 0.6373 \\ \cline{1-7} 
10-10 & \multicolumn{1}{c|}{23.2874}    & -0.0164   & \multicolumn{1}{c|}{3.2874}  & -0.0164 & \multicolumn{1}{c|}{3.9984} & 0.0560 \\ \cline{1-7} 
15-15 & \multicolumn{1}{c|}{34.1064}    & 0.1513    & \multicolumn{1}{c|}{4.1064}  & 0.1513  & \multicolumn{1}{c|}{6.2209} & 0.3445 \\ \cline{1-7} 
20-20 & \multicolumn{1}{c|}{41.7641}    & 0.0915    & \multicolumn{1}{c|}{1.7641}  & 0.0915  & \multicolumn{1}{c|}{3.7711} & 0.2466 \\ \cline{1-7} 
25-25 & \multicolumn{1}{c|}{51.7914}    & -0.0232   & \multicolumn{1}{c|}{1.7914}  & -0.0232 & \multicolumn{1}{c|}{6.1964} & 0.0325 \\ \cline{1-7} 
30-30 & \multicolumn{1}{c|}{59.5795}    & -0.0204   & \multicolumn{1}{c|}{-0.4205} & -0.0204 & \multicolumn{1}{c|}{7.5350} & 0.0374 \\ \cline{1-7} 
35-35 & \multicolumn{1}{c|}{63.9754}    & -0.0251   & \multicolumn{1}{c|}{-6.0246} & -0.0251 & \multicolumn{1}{c|}{7.0127} & 0.0036 \\ \cline{1-7} 
40-40 & \multicolumn{1}{c|}{75.6912}    & -0.0052   & \multicolumn{1}{c|}{-4.3088} & -0.0052 & \multicolumn{1}{c|}{9.7083} & 0.0068 \\ \cline{1-7} 
0-5   & \multicolumn{1}{c|}{9.1897}     & 5.6527    & \multicolumn{1}{c|}{4.1897}  & 0.6527  & \multicolumn{1}{c|}{7.3916} & 1.2036 \\ \cline{1-7} 
0-10  & \multicolumn{1}{c|}{13.1602}    & 10.1442   & \multicolumn{1}{c|}{3.1602}  & 0.1442  & \multicolumn{1}{c|}{2.1627} & 1.9077 \\ \cline{1-7} 
0-15  & \multicolumn{1}{c|}{14.5239}    & 13.5188   & \multicolumn{1}{c|}{-0.4761} & -1.4812 & \multicolumn{1}{c|}{2.6533} & 2.1334 \\ \cline{1-7} 
10-15 & \multicolumn{1}{c|}{26.9742}    & 4.5497    & \multicolumn{1}{c|}{1.9742}  & -0.4503 & \multicolumn{1}{c|}{6.8413} & 1.8721 \\ \cline{1-7} 
10-20 & \multicolumn{1}{c|}{23.2183}    & 8.4697    & \multicolumn{1}{c|}{-6.7817} & -1.5303 & \multicolumn{1}{c|}{4.7325} & 2.0071 \\ \hline
\end{tabular}
\caption{Performance in damage estimation for symmetric and asymmetric damage.}
\label{tab:nn_symmasymm}
\end{table}

\ch{To provide a metric of performance, a summarized version of the results is presented in Table \ref{tab:nn_symmasymm}, which shows the average prediction of the neural network outputs (sum and difference of the propeller cuts) for the test dataset, together with the average error and the standard deviation for both.}

For the symmetric damage flights, the difference in the cut lengths of the tips (\SI{0}{\milli\meter}) is correctly estimated by the neural network, while also showing a good performance in inferring the sum. Note that, again, the highest error is observed in the flights with the lowest degree of damage, as the behavior is not substantially affected to produce distinctive features. For higher degrees of damage, the prediction is more accurate. In the case of asymmetric damage flights, both outputs generate an accurate estimation of the sum and difference in the length, the only exception being the last flight.



For the performance in locating the damaged propeller, a confusion matrix is presented in Table \ref{tab:clasif_symasymm}, with the class number corresponding to the rotor that presents the failure (\textit{MX} corresponds to a failure in rotor \textit{X}, see Fig. \ref{fig:quad_terna} for motor numbering). While this table presents the results for the whole test dataset, some particularities must be mentioned. For example, in the flights with symmetrical damage, similar conclusions to the ones before can be drawn in the sense that, as the degree of the damage increases, the number of correctly classified samples quickly increases. As its loss of efficiency increases, so does the rotor speed, and it also greatly affects the maneuverability of the vehicle, making it easier to detect. However, for lower degrees of damage, namely the propeller with \SI{5}{\milli\meter} cuts, the classifier showed a high probability of the failure being either in the correct rotor, or in the opposite one (odd-numbered motors are opposite, so are the even-numbered ones). This issue also appears in the flights with an asymmetric damaged propeller, where the lower degrees of damage produce outputs with high probability both in the correct rotor and the opposite one.

\begin{table}[t!]
\centering
\begin{tabular}{|cc|cccl|}
\hline
\multicolumn{2}{|c|}{\multirow{2}{*}{Percentage {[}\%{]}}}                  & \multicolumn{4}{c|}{Classifier output}                                                       \\ \cline{3-6} 
\multicolumn{2}{|c|}{}                                                      & \multicolumn{1}{c|}{M1}    & \multicolumn{1}{c|}{M2}    & \multicolumn{1}{c|}{M3}    & M4    \\ \hline
\multicolumn{1}{|c|}{\multirow{4}{*}{True class}} & M1                      & \multicolumn{1}{c|}{\textbf{88.77}} & \multicolumn{1}{c|}{0.54}  & \multicolumn{1}{c|}{10.39} & 0.3   \\ \cline{2-6} 
\multicolumn{1}{|c|}{}                            & M2                      & \multicolumn{1}{c|}{0.51}  & \multicolumn{1}{c|}{\textbf{89.59}} & \multicolumn{1}{c|}{0.29}  & 9.62  \\ \cline{2-6} 
\multicolumn{1}{|c|}{}                            & M3                      & \multicolumn{1}{c|}{10.28} & \multicolumn{1}{c|}{0.74}  & \multicolumn{1}{c|}{\textbf{88.81}} & 0.18  \\ \cline{2-6} 
\multicolumn{1}{|c|}{}                            & \multicolumn{1}{l|}{M4} & \multicolumn{1}{l|}{0.44}  & \multicolumn{1}{l|}{10.29} & \multicolumn{1}{l|}{0.46}  & \textbf{88.83} \\ \hline
\end{tabular}
\caption{Performance in rotor localization for symmetric and asymmetric damage.}
\label{tab:clasif_symasymm}
\end{table}

\subsection{Baseline comparison}

\ch{To the best of our knowledge, there is no previous work that proposes an estimation of the magnitude of the failure with the same types of inputs, as this problem is generally treated as a classification one. In these cases, each specific type and magnitude of damage is treated as a different class.}
%
\ch{In order to provide an adequate baseline, while also highlighting the benefits of the proposed method, a \textit{leave one group out} experiment is performed.}

\ch{To have a point of comparison, the method proposed in \cite{Baldini2023} is used as reference, where a SVM classifier was trained to detect a finite set of three possible asymmetric damage to propellers. Here, we train a SVM classifier, considering the same quadratic kernel proposed in that work, with a tolerance of $1e-5$, using 70\%-30\% for training and test set. Training is performed only for the 13 symmetric and asymmetric damage types previously defined in Table \ref{tab:number_samples}. The features used are the same as described in the previous section, 232 values that encompass the spectral and statistical information of the sensors and command torques.}

\ch{Here, the results for one of these experiments are shown, were the type of failure left out was the 20-20. The SVM classifier was trained over the other 12 types of symmetric and asymmetric types of damage, achieving 98.07\% accuracy for the test dataset. Then, we applied this trained classifier to the 20-20 class, composed of 4204 samples (1051 samples for each of the four possible failure location). While the output could never classify these samples as the true 20-20 value, it seems to have a capability to note some similarities with other failures, as 78.54\% of the samples were classified as a 15-15 failure, a 14.96\% as a 25-25 failure, and 6.5\% as others.}

\ch{While it is expected to have zero precision in classifying a class that was never part of the training, the nature of the problem allows to define an approximation of the real failure. We define:}

\begin{align*}
    d_i = \frac{1}{N} \sum_{j=1}^{12} N_j d_{i,j}
\end{align*}

\ch{where $i\in \{ 1,2 \}$, $d_i$ is the depth of the cut of each tip, $N=4204$ is the total number of samples, $N_j$ is the number of samples predicted for class $j$, and $d_{i,j}$ is the depth of each cut in class $j$. This represents an average prediction of the damage over all the classified samples, and yields a result of $d_1=16.2635$ and $d_2 = 16.4381$, meaning a predicted damage which is almost symmetric and of around \SI{16.35}{\milli\meter} of depth.}

\begin{figure}[t!]
  \centering
  \includegraphics[width=\linewidth]{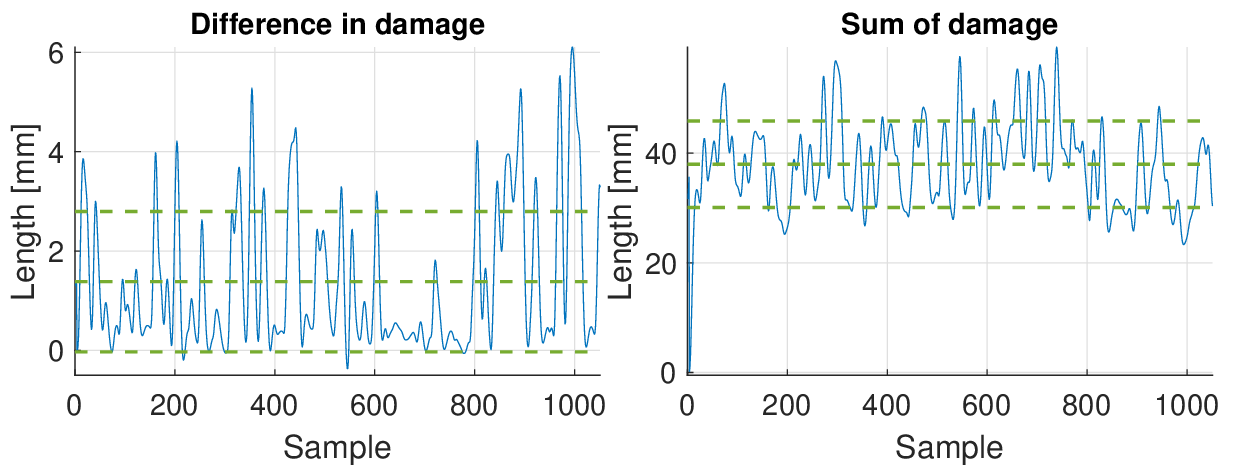}
  \caption{Estimation of the sum and difference in damage for the 20-20 damaged propeller samples.}
  \label{fig:results_nn_leaveoneout2020}
\end{figure}

\ch{To solve the same problem with the proposed neural network approach, we repeated the training of the symmetric and asymmetric damage estimation network, with exactly the same method and parameters, removing one type of failure at a time. Again, we present one the results corresponding to the case of the 20-20 propeller damage. The average predicted values of the network were \SI{37.9832}{\milli\meter} for the sum, and \SI{1.38}{\milli\meter} for the difference, which would correspond to values of $d_1=19.4527$ and $d_2 = 18.2999$. The outputs of the neural network over the full flight (where now all the samples are part of the test set) are shown in Fig. \ref{fig:results_nn_leaveoneout2020}., with the dashed green lines showing the mean and the variance of each estimation.}

To conclude, Table \ref{tab:baseline_magnitude} presents the comparison of both methods in the prediction of the damage.

\begin{table}[]
\centering
\begin{tabular}{|c|cc|cc|}
\hline
\multirow{2}{*}{Method} & \multicolumn{2}{c|}{$d_1$}          & \multicolumn{2}{c|}{$d_2$}                  \\ \cline{2-5} 
                        & \multicolumn{1}{c|}{Prediction} & Error  & \multicolumn{1}{c|}{Prediction} & Error  \\ \hline
SVM classifier \cite{Baldini2023}          & \multicolumn{1}{c|}{16.2635}    & 3.7365 & \multicolumn{1}{c|}{16.4381}    & 3.5619 \\ \hline
Neural Network (ours)          & \multicolumn{1}{c|}{19.4527}    & \textbf{0.5473} & \multicolumn{1}{c|}{18.2999}    & \textbf{1.7001} \\ \hline
\end{tabular}
\caption{Baseline comparison of the estimation of an unknown damage using a SVM classifier and a neural network in a leave-one-group-out test setting.}
\label{tab:baseline_magnitude}
\end{table}

\subsection{Longitudinal damage estimation and localization}
A similar solution is implemented for a sample that is classified as a longitudinal damage type, where the features are used to estimate the length of the cuts and locate the rotor which contains the damaged propeller. 

\begin{figure}[t!]
  \centering
  \includegraphics[width=\linewidth]{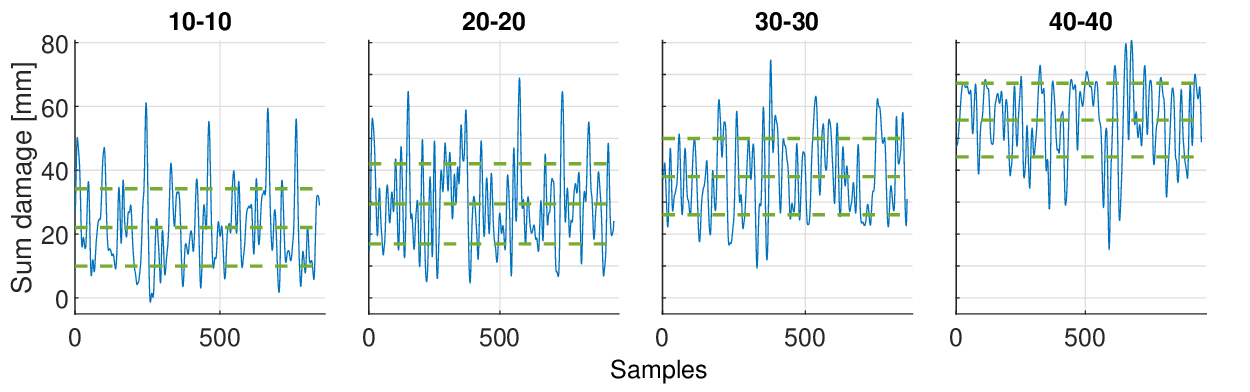}
  \caption{\ch{Estimation of the sum of damage for the longitudinal damage flights.}}
  \label{fig:results_nn_transv}
\end{figure}

\ch{In the same way as before, a prediction of the sum of the cuts over a full flight for each propeller (containing training, validation, and test samples) is shown in Fig. \ref{fig:results_nn_transv}, where the title of each sub-graph corresponds to the length cut in each of the tips, in these cases always symmetric.}
Again, the horizontal green dashed lines show the mean and plus/minus variance of the prediction for each flight.

\ch{The summarized results over the test dataset can be found in Table \ref{tab:nn_transv} presenting the average prediction of the neural network output, together with the average error and the standard deviation.}
In this case, the predictions are noisier, and the estimation deviates more from the real value as the degree of the damage increases, allegedly due to the small variations in the features for each type of damage. While the changes in the frequency spectrum are distinctive as the peak increases and moves to lower frequencies, making it easier to distinguish from the other types of damage, those changes are subtle and difficult to tell apart from each other.

\begin{table}[t!]
\center
\begin{tabular}{|cc|c|c|c|}
\hline
\multicolumn{2}{|c|}{\multirow{2}{*}{True values}} & NN output & Error    & Std. Dev. \\ \cline{3-5} 
\multicolumn{2}{|c|}{}                                     & Sum       & Sum      & Sum       \\ \hline
\multicolumn{1}{|c|}{\multirow{4}{*}{Length mean{[}mm{]}}}  & 10-10 & 21.8409   & 1.8409   & 12.1744   \\ \cline{2-5} 
\multicolumn{1}{|c|}{}                             & 20-20 & 29.2464   & -10.7536 & 12.6171   \\ \cline{2-5} 
\multicolumn{1}{|c|}{}                             & 30-30 & 37.6007   & -22.3993 & 12.2948   \\ \cline{2-5} 
\multicolumn{1}{|c|}{}                             & 40-40 & 55.2256   & -24.7744 & 12.1795   \\ \hline
\end{tabular}
\caption{Performance in damage estimation for longitudinal damage.}
\label{tab:nn_transv}
\end{table}

The confusion matrix for damage localization is presented in Table \ref{tab:clasif_transv}. In the same way as before with the symmetric and asymmetric damage, for lower degrees of damage it is more difficult to identify which rotor contains the faulty propeller. For the highest degree of damage, the classifier showed a high accuracy in detecting the correct failing rotor, and also, to a lesser extent, for the second most damaged propeller. However, for the less damaged propellers, the classifier is able to detect that there is a problem in some axis, as it consistently assigns the failure location either to the correct rotor, or to its opposite one. 

\begin{table}[t!]
\centering
\begin{tabular}{|cc|cccl|}
\hline
\multicolumn{2}{|c|}{\multirow{2}{*}{Percentage {[}\%{]}}}                  & \multicolumn{4}{c|}{Classifier output}                                                       \\ \cline{3-6} 
\multicolumn{2}{|c|}{}                                                      & \multicolumn{1}{c|}{M1}    & \multicolumn{1}{c|}{M2}    & \multicolumn{1}{c|}{M3}    & M4    \\ \hline
\multicolumn{1}{|c|}{\multirow{4}{*}{True class}} & M1                      & \multicolumn{1}{c|}{\textbf{72.42}} & \multicolumn{1}{c|}{0.30}  & \multicolumn{1}{c|}{27.00} & 0.27   \\ \cline{2-6} 
\multicolumn{1}{|c|}{}                            & M2                      & \multicolumn{1}{c|}{0.22}  & \multicolumn{1}{c|}{\textbf{72.86}} & \multicolumn{1}{c|}{0.66}  & 26.29  \\ \cline{2-6} 
\multicolumn{1}{|c|}{}                            & M3                      & \multicolumn{1}{c|}{26.89} & \multicolumn{1}{c|}{0.27}  & \multicolumn{1}{c|}{\textbf{72.64}} & 0.22  \\ \cline{2-6} 
\multicolumn{1}{|c|}{}                            & \multicolumn{1}{l|}{M4} & \multicolumn{1}{l|}{0.11}  & \multicolumn{1}{l|}{26.48} & \multicolumn{1}{l|}{0.44}  & \textbf{73.00} \\ \hline
\end{tabular}
\caption{Performance in rotor localization for longitudinal damage.}
\label{tab:clasif_transv}
\end{table}



\section{Results in outdoor environment}
\ch{As all the data previously used was collected in a controlled indoor environment, it is of interest to also evaluate the performance in an outdoor environment, where the presence of external perturbations such as wind may heavily affect the stability and maneuvers of the vehicle. 
To this end, several flights were repeated, but this time performing each flight in an outdoor setting, during a windy day. The nature of the maneuvers during the flight followed the same guidelines described in Section \ref{sec:dataacq}, where in the beginning of the flight the maneuvers performed were almost none and only to maintain hovering, then soft maneuvers to emulate normal flight, and lastly more aggressive. 
The duration of the flights was generally around \SI{120}{\second}, with some of them being shorter due to the instability of the damaged vehicle in high winds.}

\ch{The data obtained for each of the flights (accelerometers, gyroscopes, commanded torques and thrust) was again partitioned into samples of \SI{1}{\second} long (separated 32 timesteps), and then the power spectrum and other features were calculated. This samples are then fed as inputs for the previously trained neural network used for symmetric / asymmetric inference, without any kind of re-training or additional consideration for the higher perturbations.}

\begin{figure}[t!]
  \centering
  \includegraphics[width=\linewidth]{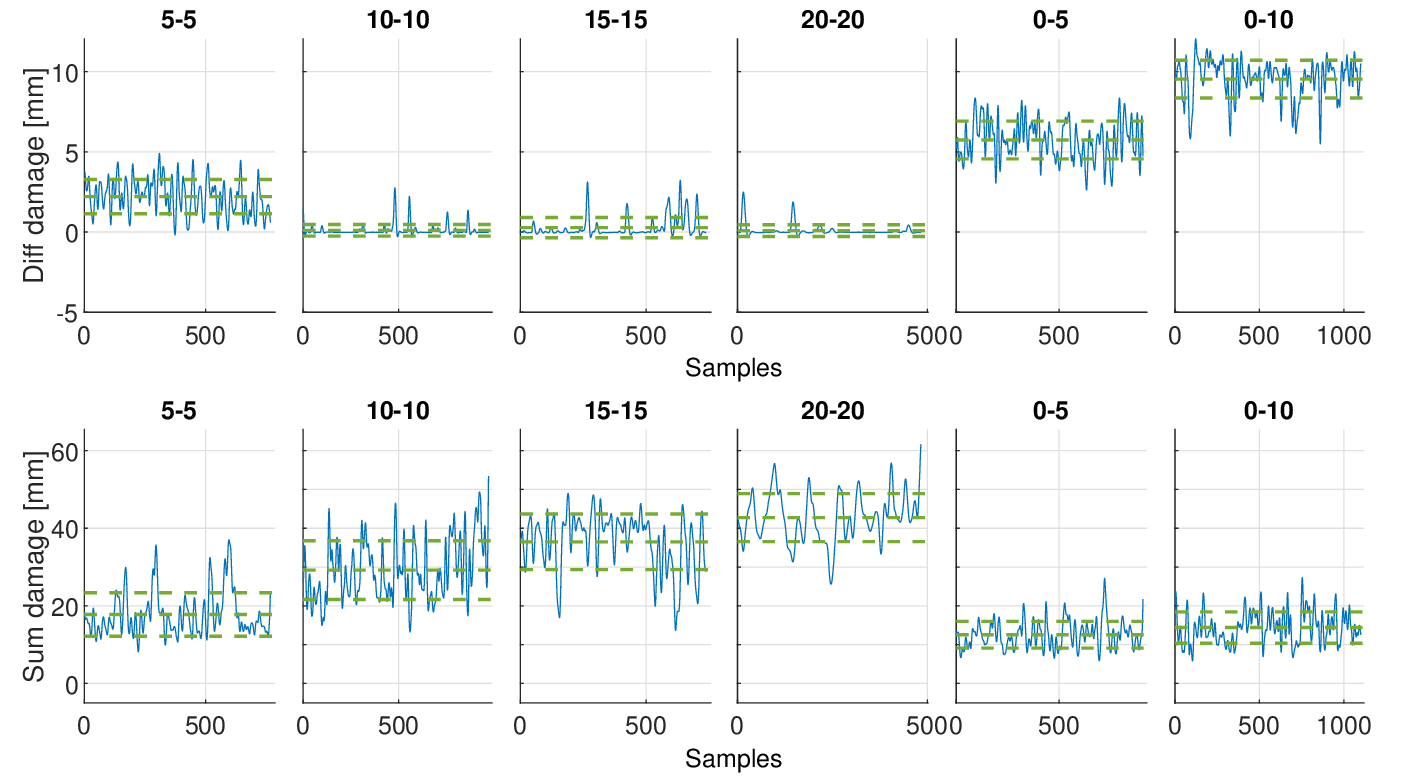}
  \caption{Estimation of the sum and difference in damage for the symmetric and asymmetric damage flights, performed in an exterior setting during a windy day.}
  \label{fig:results_nn_symmasymm_ext}
\end{figure}

\ch{The results are shown in Fig. \ref{fig:results_nn_symmasymm_ext}, in a similar way to the previous figures. Again, the first row corresponds to the predicted difference in damage between the tips for each sample, and the second row to the sum of the damage, while each column corresponds to a full flight with one of the damaged propellers. The titles of each column represent again the length cut in both tips of the propeller. The horizontal green dashed lines show the mean and plus/minus variance of the prediction for each flight.}

\ch{It can be noticed that the prediction is still highly accurate, mainly in the difference in damage (albeit there is a slight error for the 5-5 symmetric propeller). While the error in the prediction of the sum is higher for lesser damage, is still similar in magnitude to the error for the indoor flights. In consequence, the proposed method shows good adaptability to different flight conditions.}

\section{Architecture variations and considerations}\label{sec:variations}
During this work, several alternative architectures were tested to find an optimal solution for the current problem. This section presents several metrics that were used to optimize the proposed solution, and alternatives that were explored and/or discarded during the process.

\subsection{Feature importance}
To analyze which of the features provide useful information for the corresponding algorithm, the importance of the features was analyzed for each structure in the first iterations of the design. For the classifiers, the \textit{permutation\_importance} function from the \textit{sklearn} library was used, while for the neural networks, the same was done with the SHAP library for \textit{pytorch}. 
It should be noted that the latter does not yield consistently reliable results when applied to neural networks. To mitigate this and obtain a representative feature importance, several runs were made for each damage estimation network, and the results presented here show the feature importance order that was the most repeated one.

\begin{figure}[t!]
  \centering
  \includegraphics[width=1\linewidth]{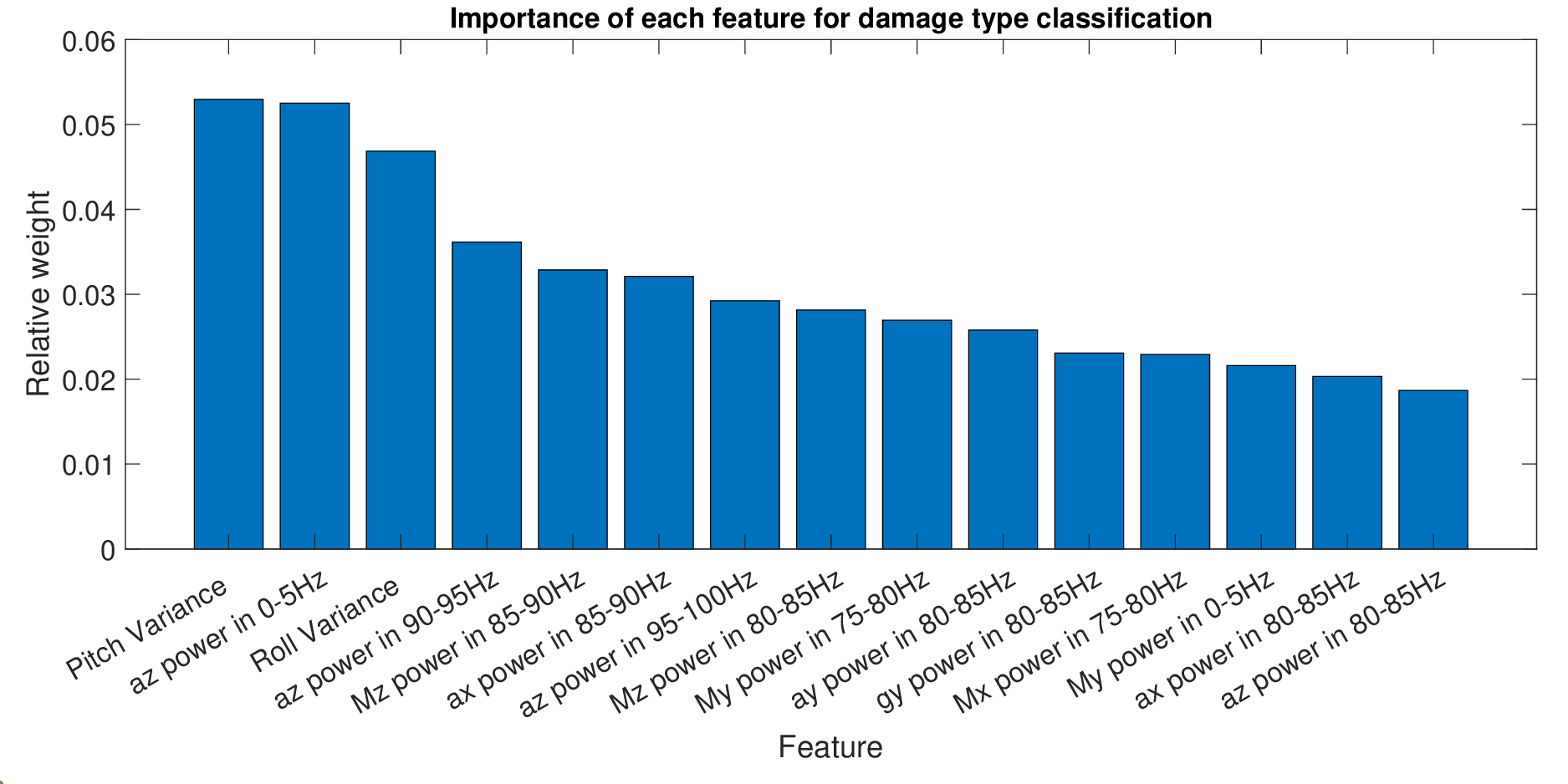}
  \caption{Feature importance for the damage type classifier.}
  \label{fig:feat_imp_faulttype}
\end{figure}

The plot in Fig. \ref{fig:feat_imp_faulttype} shows the feature importance for the damage type classifier, where the fifteen most important features are shown. As expected, the power density in some of the bands around \SI{80}{\hertz} of different inputs contribute greatly to identify the damage type. Also, the variance in the pitch and roll torque commands are two of the most important features, as small variances are an indicator of a healthy vehicle, while higher ones correspond to higher vibrations and correcting maneuvers. Surprisingly, the power density in the lowest frequencies of the $z$ accelerometer also appears as an important feature, which could be explained by more aggressive ascents and descents of the vehicle. This behavior is caused by the asymmetry in the maneuvers when one motor has lower efficiency, so the vehicle suffers a sudden loss of altitude when performing a tilt in pitch or roll.

\begin{figure}[t!]
  \centering
  \includegraphics[width=1\linewidth]{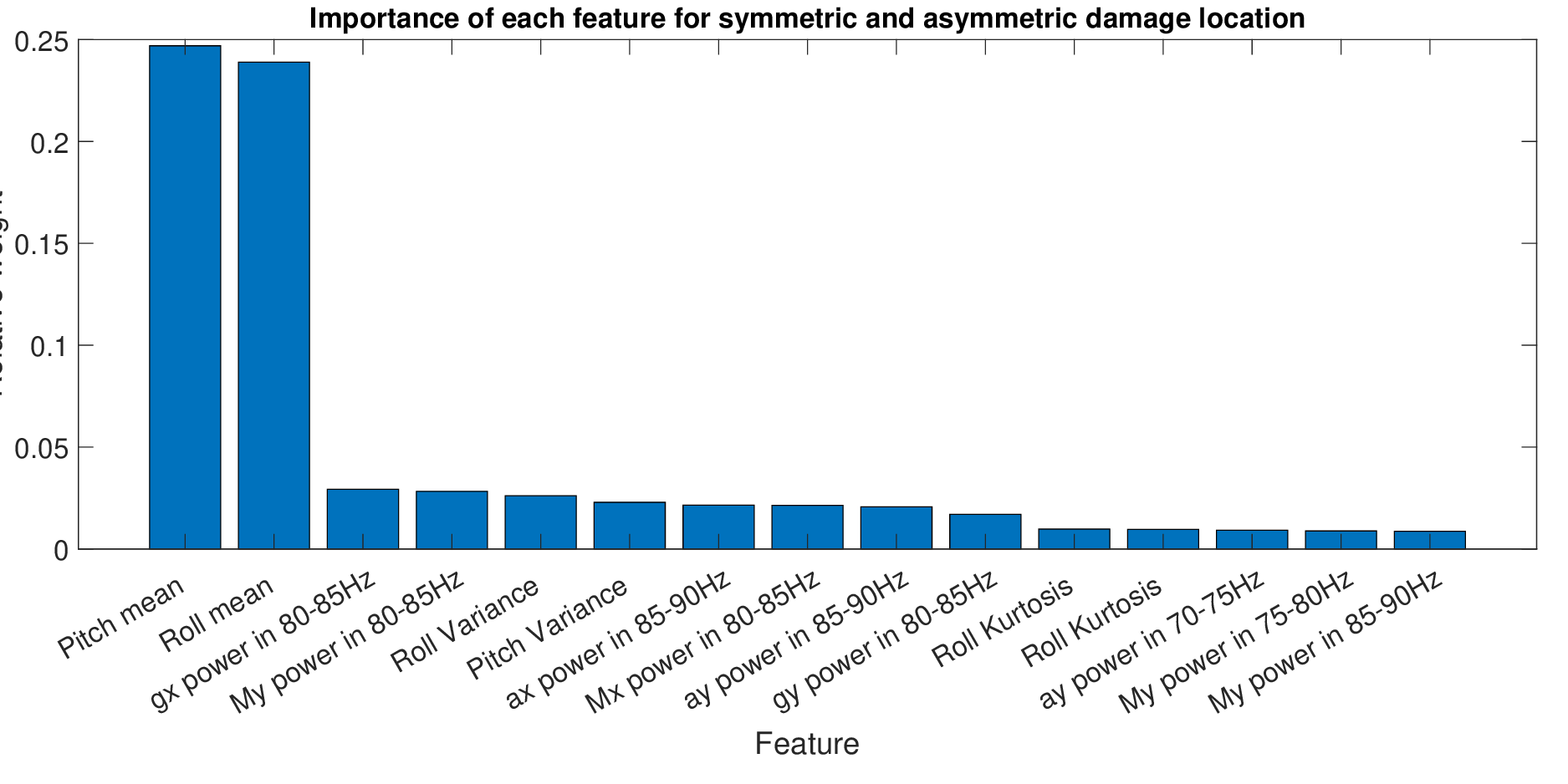}
  \caption{Feature importance for the damage localization classifier, for propellers with symmetric or asymmetric damage.}
  \label{fig:feat_imp_fsymasymm_motor}
\end{figure}

\begin{figure}[t!]
  \centering
  \includegraphics[width=1\linewidth]{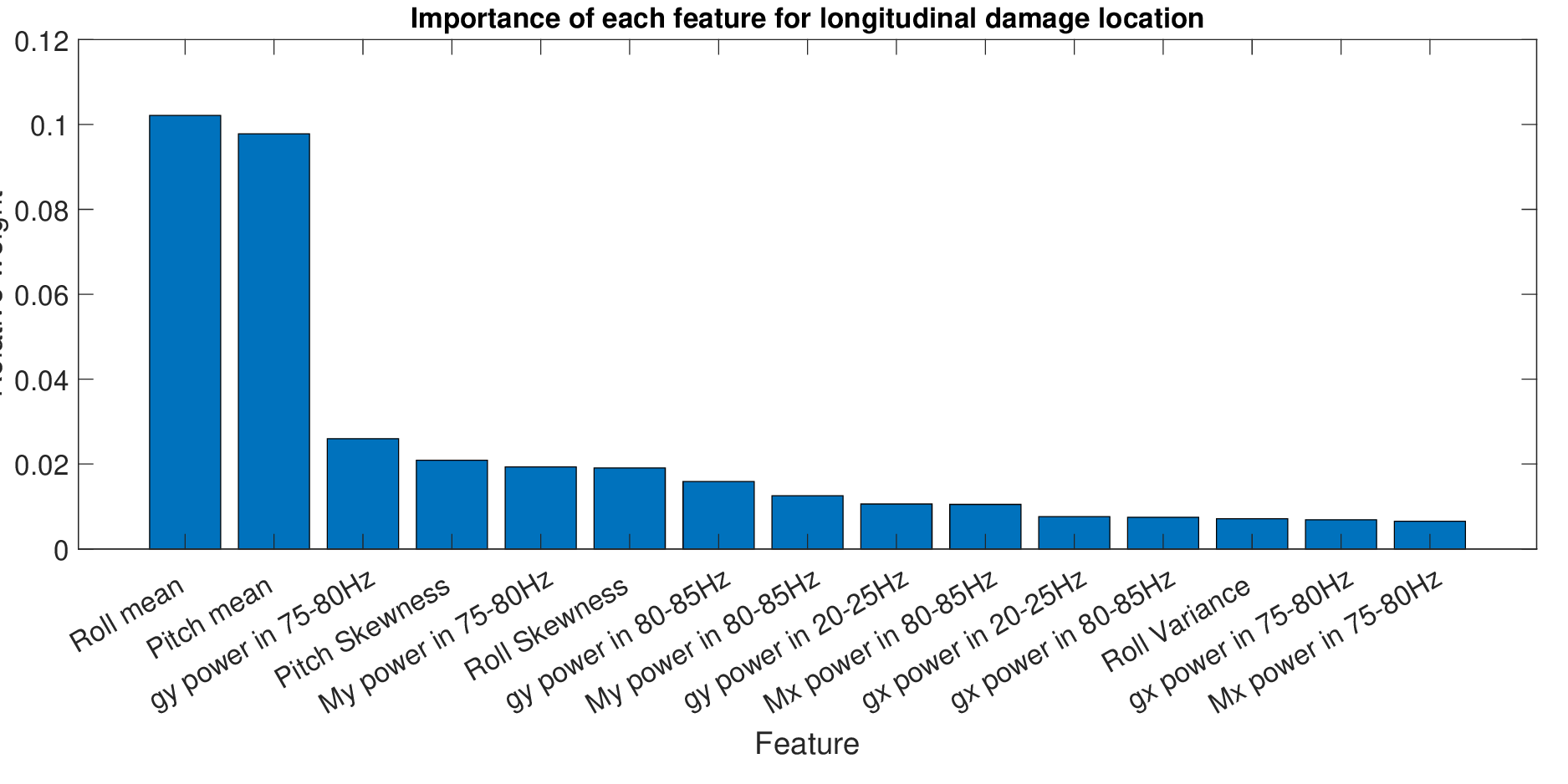}
  \caption{Feature importance for the damage localization classifier, for propellers with longitudinal damage.}
  \label{fig:feat_imp_transv_motor}
\end{figure}

For the other classifiers, Figs. \ref{fig:feat_imp_fsymasymm_motor} and \ref{fig:feat_imp_transv_motor} shows the feature importance for the damaged rotor localization. Here, the high importance of the pitch and roll mean commanded values is evident, as a constant compensation torque is needed to correct the loss of efficiency. While for the symmetric and asymmetric damage the roll and pitch commands variance appear between the most important features, for the longitudinal damage it is the skewness in those metrics that is more relevant for correct classification, which may be due to the nature of the vibrations for each propeller type. Also, as expected, the power density in some of the bands around \SI{80}{\hertz} also appear between the significant features.

\begin{figure}[t!]
  \centering
  \includegraphics[width=1\linewidth]{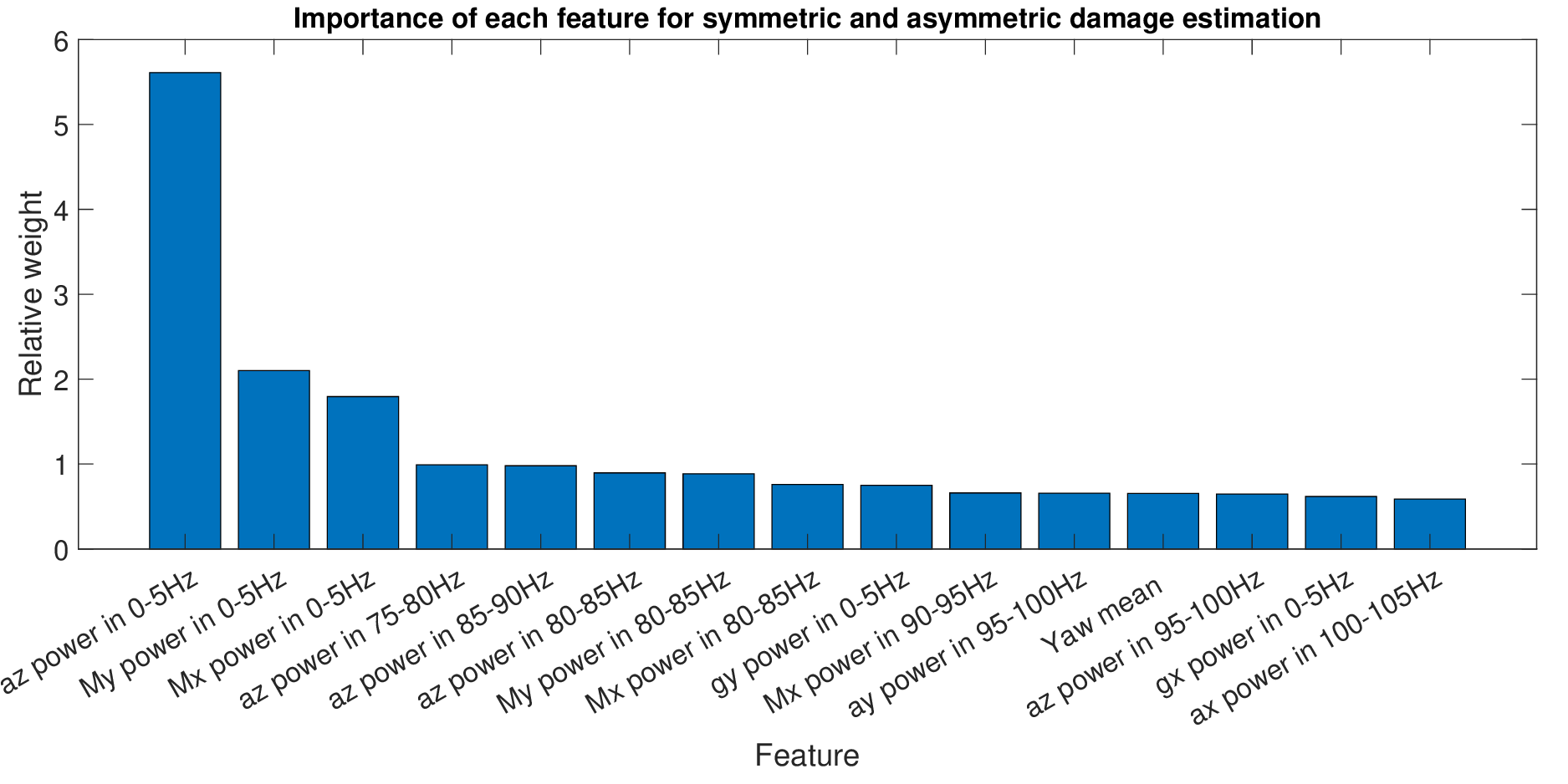}
  \caption{Feature importance for the damage estimation, for propellers with symmetric or asymmetric damage.}
  \label{fig:feat_imp_symasymm_nn}
\end{figure}

\begin{figure}[t!]
  \centering
  \includegraphics[width=1\linewidth]{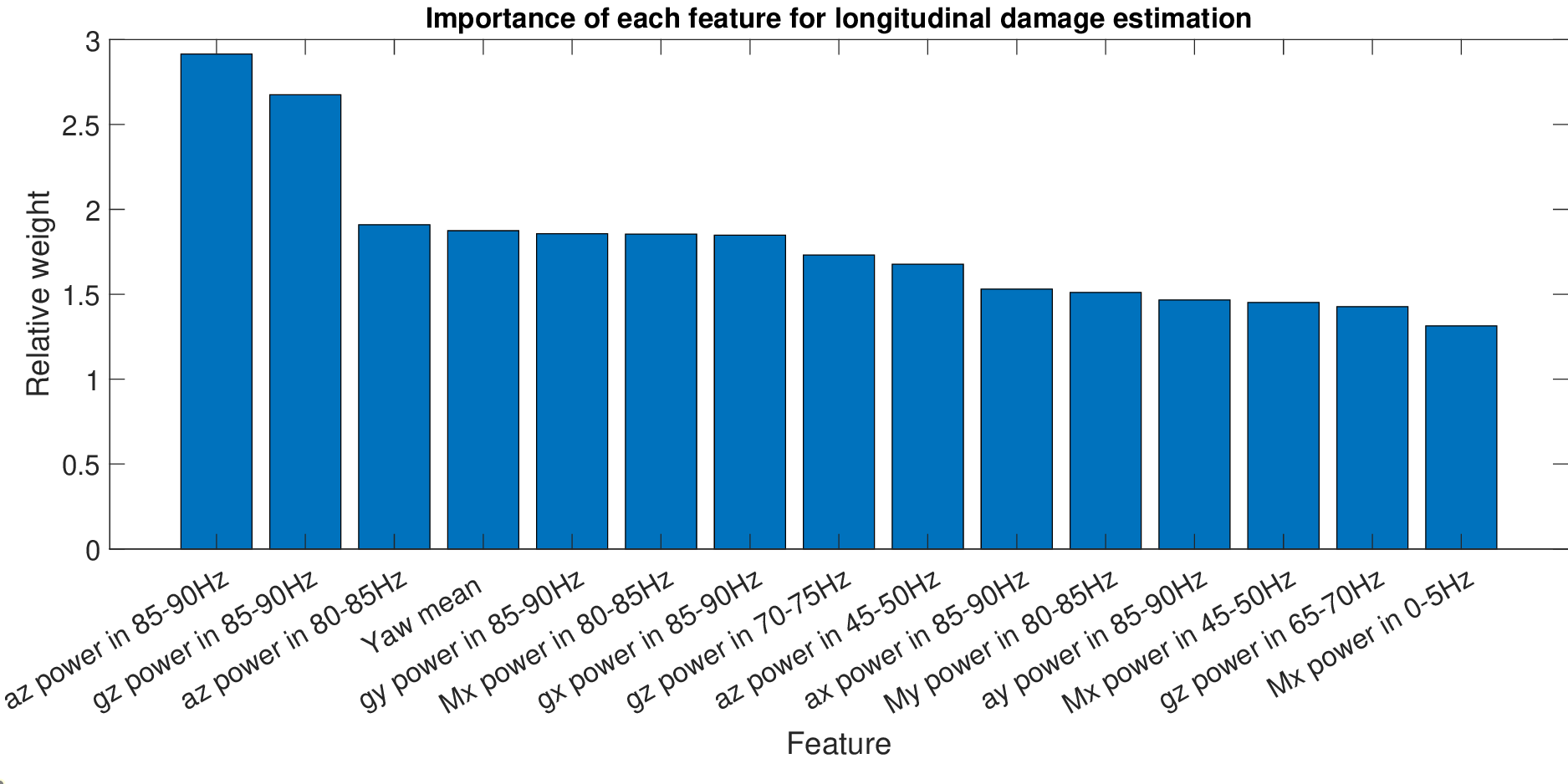}
  \caption{Feature importance for the damage estimation, for propellers with longitudinal damage.}
  \label{fig:feat_imp_transv_nn}
\end{figure}

Finally, for the neural networks, the feature importance is shown in Fig. \ref{fig:feat_imp_symasymm_nn} and \ref{fig:feat_imp_transv_nn}. For symmetric and asymmetric damage, the power density in the lowest band of the commanded torques in pitch and roll is an important feature, as it helps to estimate the loss of efficiency. Also, the power density near the \SI{80}{\hertz} bands of the $z$ accelerometer is between the important features, as it shows the same energy peaks and displacement in frequency than its $x$ and $y$ counterparts. However, during several SHAP analysis of this same network, the \SI{80}{\hertz} bands of all the accelerometers would appear in different orders of importance, with similar relative weights, so it is likely that all of them have a similar importance. A detail that is impressive is that in all SHAP runs, the low frequency energy band of the $z$ accelerometer appears as the most important feature, again probably because the vehicle suffers a sudden loss of altitude when performing a tilt in pitch or roll, and the loss is much higher as the damage in the propeller worsens.

For longitudinal damage, the most important features are generally in the frequency bands around \SI{85}{\hertz}. Going back to Fig. \ref{fig:specgram_long}, as the energy distribution for all input data suffers minimum changes along the spectrum except in the aforementioned band, it is expected for this information to be critical for correctly estimating the extent of the damage.

\subsection{Energy bands width selection}
During the feature selection phase, a study was performed to find an adequate energy band width. If the band was too wide, it would be impossible to discern whether the power density was shifting over the spectrum or not, and if it was too thin, it would greatly increase the number of features and thus the complexity of the classifiers and neural networks. So, a qualitative analysis was made, obtaining several sets of input features for different band widths, and using them to train and test the damage type classifier, the first stage in the proposed solution. 

\begin{figure}[t!]
  \centering
  \includegraphics[width=1\linewidth]{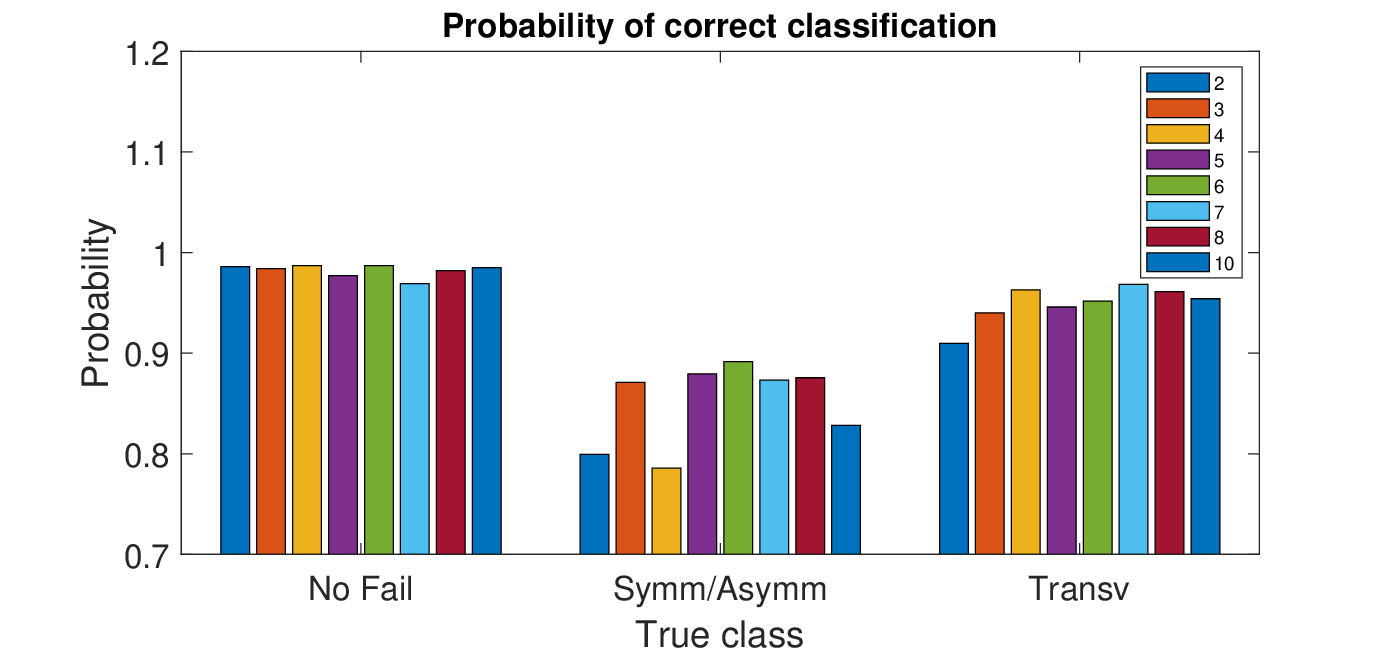}
  \caption{Probability of correct classification for the damage type classifier, for different filter widths.}
  \label{fig:filter_width_comparison}
\end{figure}

The evaluated band widths were 2, 3, 4, 5, 6, 7, 8 and \SI{10}{\hertz}, for each of them, a dataset was built consisting of samples of $n=round(111/bw)*10+12$ features, where the number 12 corresponds to the values of the mean, variance, skewness and kurtosis of the commanded torques, and the first term is the number of energy bands of the 10 sensors and torques, thus $n$ could range from 562 features for $bw=\SI{2}{Hz}$ to 122 features for $bw=\SI{10}{Hz}$. Then, one classifier for each dataset was trained after balancing the three classes using K-means, and their accuracy evaluated. Fig. \ref{fig:filter_width_comparison} shows the results of this process, where each color bar represents the accuracy of the same classifier for the three different classes. For the samples from flights which had no failures, the accuracy is high ($>$97\%) independently of the band width. For the case of the longitudinal damage, the accuracy of the classifier decreases as the band width does the same, being close to 90\% for the narrowest one. Finally, for the symmetric and asymmetric damage, there is an increase in the accuracy for the middle band widths selected, with an outlier at \SI{4}{\hertz} which cannot be explained at the moment.

The training process for the 8 classifiers shown was repeated several times to ensure consistency in the results, where the accuracy suffered minor variations over 10 different runs, even considering different number of clusters before training. Over all the different instances, the accuracy in the no failure case was indifferent of the band width, while for the longitudinal damage decreased for narrower bands, and for symmetric and asymmetric damage had an optimal around 5-\SI{6}{\hertz}. As a result, a band width of \SI{5}{\hertz} was selected.

\subsection{Ablation study}
While the SHAP analysis provides an idea of how important each feature is, from an implementation point of view it is interesting to test the robustness of the solution considering that one or more sensors, or the command signals, are not available. It is also suspected that, as the vibrations produced by the failures are present both in the sensors' and commanded torques' spectrum, there may be redundancy of data.

To provide a simple test, the symmetric and asymmetric damage estimation neural network was trained several times, each one suppressing one or more groups of features. This included eliminating either the accelerometer information, the gyroscope information, the torque information, or both the gyroscope and torque information. The training procedure was the same as previously described for the chosen solution. In order to present a simple metric for comparison, the mean square error (MSE) for the prediction was calculated both for the symmetric damage and the asymmetric damage, which is presented in Table \ref{tab:MSE_symmasymm}. Each row represents a different training instance, where the checks mark the sensors that were used, and with the first row being the proposed solution as reference. In all the new experiments, the vertical thrust was not used as feature, as it did not provide relevant features.

The table shows that the only sensor that has minor to negligible effects when not used is the set of three gyroscopes, as the MSE does not vary much w.r.t. the full feature case, it even shows a 24\% improvement in the prediction of the symmetric damage, at the expense of a 72\% increase in the MSE of the asymmetric damage. For the rest of the sensor and control signal sets, the suppression of their features cause a decrease in the performance of the neural network. If either the accelerometers or the torque commands are not used, the MSE of the symmetric damage increases between 42\% and 139\%, while the MSE of the asymmetric damage increases between 58\% and 164\%.

\begin{table}[]
\centering
\begin{tabular}{|c|c|c|c||c|c|c|c|}
\hline
Acc   & Gyro   & M   & T       & \multicolumn{1}{|p{0.8cm}|}{\centering MSE\\symm\\{[}mm{]}} & $\Delta_{MSE}$ & \multicolumn{1}{|p{0.8cm}|}{\centering MSE\\asymm\\{[}mm{]}}  & $\Delta_{MSE}$ \\ \hline
$\checkmark$  &  $\checkmark$ & $\checkmark$ & $\checkmark$ & 15.19 & -  & \textbf{2.29} & -         \\ \hline
$\checkmark$  &               & $\checkmark$ &              & \textbf{11.51} & -24\%  & 3.95 &  +72\%  \\ \hline
              & $\checkmark$  & $\checkmark$ &              & 25.53  & +68\% & 3.63 &  +58\%   \\ \hline
$\checkmark$ & $\checkmark$   &              &              & 36.30  & +139\% & 5.49 & +140\%      \\ \hline
$\checkmark$ &                &              &              & 21.50   & +42\% & 6.04  & +164\%       \\ \hline
\end{tabular}
\caption{MSE for symmetric and asymmetric damage estimation neural network, for different feature sets. Checks mark the features used in each case.}
\label{tab:MSE_symmasymm}
\end{table}

\section{Validation on the UAV-FD Dataset}

 The contributions of this paper are underscored through a validation process on the UAV-FD dataset, as introduced in \cite{Baldini2023}. This dataset, presenting a multiclass classification challenge with distinct failure rates (no failure, 5\%, and 10\%), serves as a comprehensive benchmark. Leveraging a quadratic SVM classifier, their proposed features for asymmetrical damage detection achieved remarkable success, attaining a peak accuracy of 98.5\% on the test set. This outstanding performance was realized through training on a  selected feature set of the top 51 features determined via ANOVA, encompassing crucial parameters such as motor speeds and currents, estimated and desired attitudes, velocities, and vibration data.

For our validation, we strategically narrowed down the feature set to include only gyroscope, accelerometer, and motor PWM data while applying the innovative proposed spectral energy feature set. It is noteworthy that the dataset lacked information on commanded torques; therefore, PWM data served as a surrogate, given the anticipated relationship between commanded PWM signals and torque, akin to the dynamics matrix discussed earlier in this paper. Employing a neural network as the model, as previously detailed, our approach yielded an accuracy of 98.8\% (averaged over five runs) as observed in table~\ref{tab:UAV-FD_comp}. 
This compelling result not only highlights the adaptability and effectiveness of our method but also demonstrates its robust performance on a different vehicle type, specifically a hexacopter. This validation showcases the broader applicability and reliability of our proposed methodology across varied UAV configurations, solidifying its significance in advancing the field of asymmetrical damage detection.

\begin{table}[t!]
\centering
\begin{tabular}{|c|c|c|}
\hline
Method & Accuracy \\ \hline
neural network (Ours) & \textbf{98.8\%*} \\ \hline
Quadratic SVM \cite{Baldini2023} & 98.5\% \\ \hline
Logistic regression & 94.0\% \\ \hline
\end{tabular}
\caption{\ch{Comparison of accuracies over the UAV-FD Dataset\cite{Baldini2023}. *The neural network is using a smaller feature set than the other models, using only acc, gyro and motor, and then leveraging our proposed spectral representation method.}}
\label{tab:UAV-FD_comp}
\end{table}


\section{Conclusions}
This paper aims to address the challenge of propeller fault detection in multirotors by proposing a comprehensive data-based architecture capable of detecting the existence, location, type, and degree of damage in a propeller. This is achieved utilizing only inertial and control data, which is common to any multirotor aerial vehicle and thus allows the method to be implemented in any similar platform. Moreover, it covers various types of damage, offers precise fault localization, and estimates damage magnitude.


The study builds upon previous works by incorporating commanded torques information, showcasing improvements in detection and estimation accuracy. Additionally, the usage of the sensors' energy in several frequency bands decreases the number of features without losing accuracy. The results presented in this paper signify a step forward in providing proactive maintenance solutions and in-flight fault recovery for multirotor UAVs.

The capability to extend the proposed method and its results is proved using an existing public dataset, for which the capability of fault detection is similar to that in the original study, using only the minimum necessary data. Moreover, the topology of the vehicle in the public dataset (hexarotor) is different to the one used in this manuscript (quadrotor), but still did not require any particular considerations.

As a result, this work contributes a robust data-driven architecture for propeller damage detection, localization, and characterization, with promising applications in UAV maintenance and operational reliability.

\section{Future work}
\ch{While the results are encouraging, there is still work to do to improve the capabilities and increase the range of possible applications of the proposed solution.}

\ch{Currently, the network uses the spectral information taken from \SI{1}{\second} windows, which is wide enough to avoid capturing noisy spectral data corresponding to specific maneuvers. However, the same information can be obtained from shorter or longer time windows. Using a short window would allow for quick detection times when the failure occurs during flight, but at the same time the SVM and neural networks outputs become noisier and have lower accuracy. At the moment, a good balance between the time window width and the accuracy of the solution is being sought.}

\ch{Considering that detecting failures during flight would require obtaining the spectral and statistical data, and running the classifiers and neural networks in real time, some preliminary experiments were made. Using a Raspberry Pi 4B with 4GB RAM, Python 3.9.2 and Pytorch 1.13.0, a maximum of 42.29 samples per second (\SI{23.646}{\milli\second} per sample) could be processed considering taking data from all the sensors and torques from a \SI{1}{\second} window, obtaining the spectral energy bands and mean, variance, etc., and running the damage type classifier, and the symmetric/asymmetric damage estimation neural network and damage localization classifier. As only one of the symmetric/asymmetric or longitudinal subsystem is used for a given sample (depending on the output of the failure type classifier), this experiment provides encouraging results, considering that no code optimization was made. Even further, most of the processing time (\SI{22.556}{\milli\second} - around 95\%) is used for pre-processing to obtain the energy bands, task which is easily improved by using better implementations, or even aproxiamtions such as the Welch method. It is also planned to implement the solutions in a Jetson TX4 currently mounted in one of the vehicles to obtain even faster results.}



\bibliographystyle{IEEEtran}
\bibliography{biblio.bib}

\end{document}